\documentclass[12pt]{iopart}

\usepackage{graphicx}
\usepackage{subfigure}
\usepackage{color}
\usepackage{balance}
\usepackage{lmodern}
\usepackage{bm}
\usepackage{amsfonts}
\usepackage{algpseudocode}
\usepackage{algorithm}
\usepackage{hyperref}
\usepackage{amsopn}
\DeclareMathOperator*{\argmin}{argmin}
\DeclareMathOperator*{\argmax}{argmax}

\begin{document}
\nocite{*}
\title[Direct high-order edge-preserving regularization for CT]{Direct high-order edge-preserving regularization for tomographic image reconstruction}

\author{Daniil~Kazantsev$^{1,2}$, Evgueni~Ovtchinnikov$^3$, William~R. B. Lionheart$^5$, Philip~J. Withers$^{1,2}$, Peter~D. Lee$^{1,2}$}

\address{$^1$ The Manchester X-Ray Imaging Facility, School of Materials, The University of Manchester, Manchester, M13 9PL, UK}
\address{$^2$ The Manchester X-Ray Imaging Facility, Research Complex at Harwell, Didcot, Oxfordshire, OX11 0FA, UK}
\address{$^3$ Visualization Group, Research Complex at Harwell, STFC, Oxfordshire, OX11 0QX, UK}
\address{$^4$ School of Mathematics, Alan Turing Building, The University of Manchester, M13 9PL, UK}

\ead{daniil.kazantsev@manchester.ac.uk}

\maketitle

\begin{abstract}
In this paper we present a new two-level iterative algorithm for tomographic image reconstruction. The algorithm uses a regularization technique, which we call edge-preserving Laplacian, that preserves sharp edges between objects while damping spurious oscillations in the areas where the reconstructed image is smooth. Our numerical simulations demonstrate that the proposed method outperforms total variation (TV) regularization and it is competitive with the combined TV-$\ell_{2}$ penalty. Obtained reconstructed images show increased signal-to-noise ratio and visually appealing structural features. Computer implementation and parameter control of the proposed technique is straightforward, which increases the feasibility of it across many tomographic applications. In this paper, we applied our method to the under-sampled computed tomography (CT) projection data and also considered a case of reconstruction in emission tomography The MATLAB code is provided to support obtained results. 
\end{abstract}

\section{Introduction} \label{sec:intro}
Frequently, in X-ray computed tomography (CT), the amount of collected projection data is lower that it is required by the Nyquist sampling theorem \cite{kak}. In medical imaging, the restrictions are applied to minimize the ionizing radiation which can harm living tissue cells \cite{wernick}. In material science, the aim is to better resolve temporal resolution via higher frame rate acquisition \cite{kazantsev1}. In such cases of limited data, iterative techniques can provide better reconstructions than analytical methods \cite{qi}. 

Dealing with ill-posed and ill-conditioned inverse problems, iterative methods require regularization to constrain the space of desirable solutions. Due to edge-preserving properties, total variation (TV) regularization \cite{rudin} has been extensively used in tomographic iterative reconstruction for the last three decades. However, TV penalty produces piecewise-constant images (so-called ``cartoon'' or ``staircasing'' effect) even if the original object is smooth \cite{wang}. This ``cartoon'' effect can be particularly undesirable in emission tomography (ET) \cite{wernick}, such as Positron Emission Tomography (PET) or Single-Photon Emission Computed Tomography (SPECT) where due to limited resolution of the imaging system and low photon-number statistics, reconstructed images are naturally blurred. Since in ET the activity distribution is piecewise-smooth and the use of TV penalty can result in undesirable artifacts, this can potentially bias the following clinical interpretation of reconstructed images \cite{mumcuoglu}. 

Various improvements have been made to reduce ``cartoon'' appearance of the recovered images \cite{chan1,chan2,bredies,liu,papafitsoros,do,lysaker}. The most straightforward way is to couple TV term with $\ell_{2}$ norm \cite{chan1,chan2,papafitsoros,do}. The goal of this approach is to establish a trade-off (by means of regularization constants) between piecewise-constant and piecewise-smooth regularization terms. This, however, complicates the optimization problem and for practical reconstruction  purposes it can be a challenging task to establish the required regularization parameters. Therefore, the use of a single penalty term which can accommodate the required properties is highly desirable for tomographic reconstruction. Some single term penalties have been proposed for image denoising and they based on the \textit{edge preserving Laplacian} \cite{liu,lysaker} or generalized forms of TV norm \cite{bredies}. 

In this paper, we propose a high-order penalty which has similarities with the Laplacian based methods \cite{liu,lysaker} but it also possess some of unique qualities. We adapt our high-order penalty to the tomographic reconstruction problem using the two-step fixed point iteration algorithm \cite{vogel}. To demonstrate applicability of the proposed regularizer to image reconstruction  problems we compare it with the classical TV \cite{rudin} and combined TV-$\ell_{2}$ methods for CT and ET image reconstruction. All techniques are compared quantitatively and visually.

\section{Method}\label{sec:method}
\subsection{Image reconstruction problem}
Tomographic image reconstruction problem consists of determining the inner structure of an object based on its X-ray observations from the several different angular positions. Incoming photons with different energies (due to absorption) are registered by detectors and information about the path length a photon has travelled along the line can be decoded \cite{kak}. Therefore, by solving the inverse problem where projection data is given, the degree of absorption or attenuation coefficient of the object can be recovered. In mathematical terms, the reconstruction problem can be formulated as the least squares (LS) problem:
\begin{eqnarray}
\label{ls}
\hat {\bm{u}} =  \argmin\limits_{\bm{u}} \left\Vert \mathrm{A} \bm{u} - \bm{b} \right\Vert_{2}^2,
\end{eqnarray}
where $\bm{b} \in \mathbb{R}^{M}$ is a discrete function of the number of the detector bins and the observation angles describing the projection data (sinogram), $\bm{u} \in \mathbb{R}^{N}$ is a function of spacial variables describing the observed object (e.g. density of the object's material) and $\mathrm{A}: \mathbb{R}^{{N}}\rightarrow  \mathbb{R}^{{M}}$ is a sparse system projection matrix (discrete approximation of the continuous Radon transform \cite{kak}) mapping the ``space of objects'' to the ``space of observations''. In continuous space  the operator $\mathrm{A}$ is an integral operator, and zero is a condensation point of its singular values, which makes the problem (\ref{ls}) ill-posed \cite{hansen}. 

Depending on the numbers of spacial grid points, detectors and angles, the matrix $\mathrm{A}$ can be ``fat'' (the number of rows is less than the number of columns) or ``tall'' (the number of rows is greater than the number of columns). Irrespective of the shape, the singular values of $\mathrm{A}$ form a very tight cluster near zero owing to the same property of the integral operator from which $\mathrm{A}$ derives \cite{hansen}.

\subsection{Regularization}
The quadratic functional $J(\bm{u})=\left\Vert \mathrm{A} \bm{u} - \bm{b} \right\Vert_{2}^2$ with its ``normal'' form $\mathrm{A}^{*}\mathrm{A}\bm{u} = \mathrm{A}^{*}\bm{b}$ can be minimized by a suitable iterative minimization algorithm,  e.g. Conjugate Gradient Least Squares (CGLS) algorithm \cite{nocedal}, 
however, the convergence of iterations can be very slow
because of the poor conditioning of the Hessian
$\mathrm{A}^{*}\mathrm{A}$ \cite{vogel}. The slow convergence of iterations is closely related to another difficulty in dealing with this kind of problem,
which is the ill-posedness of the problem.
Generally, the solution $\hat {\bm{u}}$ of the problem 
(\ref{ls}) is not unique (if $\mathrm{A}$ is ``fat''),
and even if it is, in practical computation $\hat {\bm{u}}$ is indistinguishable from any $\hat {\bm{u}} + \bm{h}$ if $\left\Vert \mathrm{A} \bm{h}\right\Vert_{2}$ is below the round-off error level, which, in the case at hand, may happen even if $\left\Vert \bm{h}\right\Vert_{2}$ is substantial.

Both difficulties can be tackled by a \textit{regularization} technique, whereby (\ref{ls}) is replaced with 
\begin{eqnarray}
\label{ls.reg}
\hat {\bm{u}}_\alpha = \argmin\limits_{\bm{u}} \psi_\alpha(\bm{u}), 
\quad
\psi_\alpha(\bm{u}) = \frac{1}{2}\left\Vert \mathrm{A} \bm{u} - \bm{b}  \right\Vert_{2}^2 + \alpha R(\bm{u}),
\end{eqnarray}
where $R(\bm{u})$ is a suitable regularization functional, and $\alpha$ is a positive scalar parameter. The LS data term in (\ref{ls.reg}) can be replaced by other fit-to-data functional, for example, Poisson log- likelihood functional, which is better suited noise model for ET reconstruction \cite{qi}. Optimization problem with regularized Poisson log-likelihood is
$$
\hat{\bm{u}}_{\alpha, P} = \argmax\limits_{\bm{u}} \psi_{\alpha, P}(\bm{b}, \mathrm{A}\bm{u}), 
$$
\begin{eqnarray}
\label{kl.reg}
\psi_{\alpha, P}(\bm{b}, \mathrm{A}\bm{u}) = \sum_{i}^{M}\{-[\mathrm{A}\bm{u}]_{i} + \bm{b}_{i}\log[\mathrm{A}\bm{u}]_{i}\} + \alpha R(\bm{u}).
\end{eqnarray}
Now we consider various regularization terms $R(\bm{u})$ that can be used in (\ref{ls.reg}) and (\ref{kl.reg}) cost functions. A prime example of the regularization is known as Tikhonov's \cite{vogel}, where 
\begin{eqnarray}
\label{Tikhpen}
R_{\ell_{2}}(\bm{u}) = \left\Vert\bm{u}\right\Vert_{2}^2.
\end{eqnarray}
Regularization makes the minimization problems (\ref{ls.reg})  and (\ref{kl.reg}) well-posed, the eigenvalues of the Hessian of $\psi_\alpha(\bm{u})$ being not less than $\alpha$. Tikhonov regularization is quadratic, therefore high frequencies which are related to the object boundaries are penalized more, resulting in a smooth recovery of $\hat {\bm{u}}$. To preserve boundaries one needs to consider non-quadratic penalties, e.g. the TV penalty \cite{rudin} can significantly improve oscillatory solutions. The differentiable (due to small $\epsilon$ constant) TV penalty is given as: 
\begin{eqnarray}
\label{TVpen}
R_{TV}(\bm{u}) = \left\Vert|\nabla \bm{u}|_{\epsilon}\right\Vert_{1}= \left\Vert\sqrt{\bm{u}_{x}^2 + \bm{u}_{y}^2 + \epsilon^2}\right\Vert_{1}.
\end{eqnarray}
Unlike Tikhonov's penalty (\ref{Tikhpen}), TV can recover function while preserving sharp discontinuities, however, it also leads to the patchy appearance of the image \cite{chan1,chan2,bredies,liu,papafitsoros,do,lysaker}. Since TV is unable to recover smooth variations alone, it can be coupled with $\ell_{2}$-norm based penalty \cite{chan1,chan2,papafitsoros,do} resulting in the combined functional
\begin{eqnarray}
\label{combpen}
R_{TV-\ell_{2}}(\bm{u}) =  \left\Vert|\nabla \bm{u}|_{\epsilon}\right\Vert_{1} + \frac{\mu}{\alpha} \left\Vert\bm{u}\right\Vert_{2}^2,
\end{eqnarray} 
where $\mu$ is an additional regularization parameter for the quadratic term. In  \cite{chan1}, the following combined functional has been proposed:
\begin{eqnarray}
\label{combpen2}
R_{TV-\ell_{2}}(\bm{u}) = \left\Vert|\nabla \bm{u}|_{\epsilon}\right\Vert_{1} + \frac{\mu}{\alpha} \left\Vert \frac{(\Delta \bm{u})^{2}}{|\nabla\bm{u}|^{3}}\right\Vert_{2},
\end{eqnarray}  
where $\Delta \bm{u} = \bm{u}_{xx} + \bm{u}_{yy}$ denotes an isotropic Laplacian. In this work, we will be using (\ref{TVpen}) and (\ref{combpen2}) functionals for comparison with the proposed penalty. 

\subsection{A note on the regularization error}

It is intuitively obvious that the regularization parameter $\alpha$ in (\ref{ls.reg}) and (\ref{kl.reg}) must not be large. In order to get some further insight into the issue, let us estimate the difference between $\hat{\bm{u}}_\alpha$ and $\hat{\bm{u}}$ assuming for simplicity that all singular values of $\mathrm{A}$ are positive.

Let us assume $R(\bm{u}) = \|\mathrm{R} \bm{u}\|^2_{2}$, where $\mathrm{R}$ is a square non-degenerate matrix, and denote $\mathrm{M} = \mathrm{A}^* \mathrm{A}$, and $\mathrm{N} = \mathrm{R}^* \mathrm{R}$.
In the case at hand, $\hat{\bm{u}}_\alpha$ (where $\alpha$ may be zero)
satisfies
\begin{eqnarray}
\label{ls.reg.eq}
(\mathrm{M} + \alpha \mathrm{N})\hat{\bm{u}}_\alpha = \mathrm{A}^* \bm{b},
\end{eqnarray}
which implies the following equation for the regularization error,
$\bm{h}_\alpha = \hat{\bm{u}}_\alpha - \hat{\bm{u}}$:
\begin{eqnarray}
\label{corr}
(\mathrm{M} + \alpha \mathrm{N})\bm{h}_\alpha = -\alpha \mathrm{N} \hat{\bm{u}}.
\end{eqnarray}
Multiplication by $\mathrm{M}^{-1} \mathrm{N} \bm{h}_\alpha$ yields
\begin{eqnarray}
\label{corr.prod}
((\mathrm{N} + \alpha \mathrm{N} \mathrm{M}^{-1} \mathrm{N})\bm{h}_\alpha, \bm{h}_\alpha) = 
-\alpha (\mathrm{N} \hat{\bm{u}}, \mathrm{M}^{-1} \mathrm{N} \bm{h}_\alpha).
\end{eqnarray}
Now, in the left-hand side of (\ref{corr.prod}) we have
\begin{eqnarray*}
((\mathrm{N} + \alpha \mathrm{N} \mathrm{M}^{-1} \mathrm{N})\bm{h}_\alpha, \bm{h}_\alpha)
\ge
(\mathrm{N} \bm{h}_\alpha, \bm{h}_\alpha) = 
(\mathrm{R}^* \mathrm{R} \bm{h}_\alpha, \bm{h}_\alpha) = 
\left\Vert|\mathrm{R} \bm{h}_\alpha\right\Vert_{2}^2,
\end{eqnarray*}
and in the right-hand side
$(\mathrm{N} \hat{\bm{u}}, \mathrm{M}^{-1} \mathrm{N} \bm{h}_\alpha) =$
\begin{eqnarray*}
(\mathrm{N} \mathrm{M}^{-1} \mathrm{N} \hat{\bm{u}}, \bm{h}_\alpha) =
(\mathrm{R} \mathrm{M}^{-1} \mathrm{N} \hat{\bm{u}}, \mathrm{R} \bm{h}_\alpha) \le
\left\Vert\mathrm{R} \mathrm{M}^{-1} \mathrm{N} \hat{\bm{u}}\right\Vert_{2} \left\Vert\mathrm{R} \bm{h}_\alpha\right\Vert_{2}.
\end{eqnarray*}
Thus, (\ref{corr.prod}) implies
\begin{eqnarray}
\label{reg.err}
\|\mathrm{R} \bm{h}_\alpha\|_{2} \le \alpha \left\Vert\mathrm{R} \mathrm{M}^{-1} \mathrm{N} \hat{\bm{u}}\right\Vert_{2}.
\end{eqnarray}

\subsection{Edge-preserving piecewise-smooth regularization}

At first glance, the regularization appears to be merely a compromise move that distorts the problem so that it becomes solvable. While this is certainly so in the case of Tikhonov's regularization, an alternative viewpoint can be offered,
which is helpful in designing a proper regularization for the problem at hand. One observes that the problem (\ref{ls}) can only be solved approximately, if only because of the inexactness of computer arithmetic. One observes further that it may have infinitely many approximate solutions that are indistinguishable in practical computation, as pointed out in the previous section, if one is only guided by the smallness of the goal data fidelity functional $\left\Vert\mathrm{A} \bm{u} - \bm{b}\right\Vert_{2}^{2}$. The regularization can be viewed as some kind of additional criterion that helps to verify whether a particular computed solution is acceptable. This viewpoint is supported by the fact that in the case where $R(u) = \left\Vert\mathrm{R}\bm{u}\right\Vert_{2}^2$, the regularized problem (\ref{ls.reg}) is equivalent to the original problem (\ref{ls}) for these extended $\mathrm{A}$ and $\bm{b}$:
\begin{eqnarray}
\label{ls.ext}
\mathrm{A}_\alpha = \left[
\begin{array}{c}
\mathrm{A} \\
\alpha R
\end{array}
\right],
\quad
\bm{b}_\alpha = \left[
\begin{array}{c}
\bm{b} \\
0
\end{array}
\right].
\end{eqnarray}
In the problem (\ref{ls}), $\mathrm{A}$ is a discretization of an integral operator, owing to which $\left\Vert\mathrm{A} \bm{h}\right\Vert_{2}$ is small on oscillating grid functions $\bm{h}$ with wavelengths that are close to the grid step. Hence, if one directly applies e.g. CGLS algorithm to the minimization of the quadratic functional (\ref{ls}), then after sufficiently many iterations one is likely to end up with a quickly oscillating approximate solution $\hat{\bm{u}}$. But most images that one encounters in practice do not feature such oscillations and can be actually represented by piecewise-smooth functions $\hat{\bm{u}}$. This suggests that the value of $R(\bm{u})$ should be large on short-wavelength functions $\bm{u}$. At the same time, $R(\bm{u})$ should remain small on the boundaries (walls) between objects constituting the image, where $\bm{u}$ is discontinuous or has large gradients.

The following regularizer is therefore suggested:
\begin{eqnarray}
\label{wcr}
R_{EL}(\bm{u}) = 
\left\Vert \bm{w_x} \frac{\partial^2 \bm{u}}{\partial x^2} \right\Vert_{2}^2
+
\left\Vert \bm{w_y} \frac{\partial^2 \bm{u}}{\partial y^2} \right\Vert_{2}^2,
\end{eqnarray}
where the weights $\bm{w_x}$ and $\bm{w_y}$ are given by
\begin{eqnarray}
\label{wcr.w}
\bm{w_x} = 
\left(
1 + \beta
\left(
\frac{1}{a_x}
\frac{\partial \bm{u}}{\partial x}
\right)^2
\right)^{-1},
\quad
\bm{w_y} = 
\left(
1 + \beta
\left(
\frac{1}{a_y}
\frac{\partial \bm{u}}{\partial y}\right
)^2
\right)^{-1},
\end{eqnarray}
$\beta$ is a positive scalar parameter and it determines which points are considered to be on an edge between objects rather than inside it - the smaller the value of $\beta$, the smaller the edge area. Constants $a_x$ and $a_y$  are the average
$x$- and $y$-derivatives of $\bm{u}$. These averages are
introduced purely for the sake of scale-invariance, and can be computed e.g. as $a_x = 2u_{max}/d_x$ and $a_y = 2u_{max}/d_y$
where $u_{max}$ is the maximum of $\bm{u}$ and $d_x$ and $d_y$ are the sizes of the square containing the image. 

By design, $R_{EL}(\bm{u})$ (where EL stands for the Edge preserving Laplacian) is large on a short-wavelength functions $\bm{u}$ with small oscillation amplitude (approaching to isotropic smoothing) inside the reconstructed objects, and small on the edges between them (owing to the smallness of weights $\bm{w_x}$ and $\bm{w_y}$). The edge points are identified as points where the derivatives of $\bm{u}$ are significantly higher than their average values.

The regularizer (\ref{wcr}) shares some similarities with the high-order penalty proposed for image denoising in \cite{liu}:
\begin{eqnarray}
\label{liumeth}
\hat{u} = \argmin\limits_{u} \int_{\Omega} \omega |\Delta u|d\bm{\mathrm{x}} + \frac{\lambda}{2}\int_{\Omega} (u - u_{0})^{2}d\bm{\mathrm{x}},
\end{eqnarray}
where $u_{0}$ is a noisy image and weights defined as $\omega =  \left(1 + \beta|\nabla G_{\sigma} \ast u_{0} |\right)^{-1},$ where $G_{\sigma}$ denotes the Gaussian filter with the kernel width $\sigma$ and $\ast$ is the convolution. There are few differences of the proposed penalty (\ref{wcr}) from (\ref{liumeth}). For tomographic reconstruction a good initialization $\bm{u_{0}}$ is not usually available, therefore our regularizer is built on a different principle of ``catching'' prominent edges. The proposed penalty (\ref{wcr}) is independent of the smoothed gradient while remain stable to the  presence of noise (we will justify it with our numerical experiments). Moreover, we incorporate directional high-order smoothing in our term with consideration of variant weights (second derivatives are weighted \textit{independently}), in our penalty $\bm{w_x} \neq \bm{w_y}$ (cross partial derivatives can be also added). Overall the penalty (\ref{wcr}) is more rigorous than (\ref{liumeth}) and it will not allow any harmonic noise into the solution. 


\subsection{Regularized reconstruction with LS data fidelity term}
In order to simplify the computation of the gradient of $\psi_\alpha(\bm{u})$ and dealing with the minimization of a quadratic functional, we resort to an inner-outer iterative scheme with lagged diffusivity fixed point iteration \cite{vogel}, with $\bm{w_x}$ and $\bm{w_y}$ only updated on restarts (this approach helps to deal with non-convexity of penalty (\ref{wcr}) cf. widely used trust region technique \cite{nocedal}). 

Let us denote by $\mathrm{L}_x$ and $\mathrm{L}_y$ the matrices representing the discretized second partial $x$- and $y$-derivatives with Neumann boundary conditions, and by $\mathrm{W}_x$ and $\mathrm{W}_y$ the diagonal matrices representing $\bm{w_x}$ and $\bm{w_y}$. The symmetric gradient matrix $\mathrm{R}(\bm{u})$, such as $\nabla R(\bm{u}) = \mathrm{R}(\bm{u}) \bm{u}$, is given for EL term as (ignoring the dependence of $\bm{w_x}$ and $\bm{w_y}$ on $\bm{u}$ -- cf. inner-outer iterative scheme)
\begin{eqnarray}
\label{wcr.H}
\mathrm{R}_{EL} = \mathrm{L}_x^* \mathrm{W}_x^* \mathrm{W}_x \mathrm{L}_x + \mathrm{L}_y^* \mathrm{W}_y^* \mathrm{W}_y \mathrm{L}_y.
\end{eqnarray}
We compare the proposed EL penalty with TV regularization  term (\ref{TVpen}) with the following gradient matrix $\mathrm{R}_{TV}$:
\begin{eqnarray}
\label{wcr.HTV}
\mathrm{R}_{TV} = \mathrm{D}_{x}^* \Phi(\bm{u}) \mathrm{D}_{x} + \mathrm{D}_{y}^* \Phi(\bm{u}) \mathrm{D}_{y},
\end{eqnarray}
where $\mathrm{D}_{x}$ and $\mathrm{D}_{y}$ are matrices representing the discretized first partial $x$- and $y$-derivatives  with Neumann boundary conditions, $\Phi(\bm{u}) = \mathit{diag}(\phi'(\bm{u}))$ is a diagonal matrix whose diagonal elements are $\phi'(\bm{u})$, $\phi(t) = 2\sqrt{t + \epsilon^2}$.  Different choices can be used for $\phi(t)$ function to approximate $\ell_{1}$ norm, for example the Huber function \cite{vogel,kazantsev}.

We also compare the proposed penalty (\ref{wcr}) with TV-$\ell_{2}$ functional with the gradient matrix defined as
\begin{eqnarray}
\label{wcr.HTVL2}
 \mathrm{R}_{TV-\ell_{2}} = \mathrm{D}_{x}^* \Psi(\bm{u}) \mathrm{D}_{x} + \mathrm{D}_{y}^* \Psi(\bm{u}) \mathrm{D}_{y}  + \mathrm{L}_{x}^{*} \Upsilon(\bm{u}) \mathrm{L}_{x} + \mathrm{L}_{y}^{*} \Upsilon(\bm{u}) \mathrm{L}_{y},
\end{eqnarray}
where $\Psi(\bm{u})$ is a diagonal matrix with elements $\alpha/|\nabla \bm{u}|_{\epsilon}$ and $\Upsilon(\bm{u})$ is a diagonal matrix with elements $2\mu/|\nabla \bm{u}|^{3}_{\gamma}$. Values $\epsilon$ and $\gamma$ were chosen similar to \cite{chan1}, $\epsilon = 10^{-5}u_{max}$ and $\gamma = u_{max}^{2}$.

Here we used the lagged diffusivity fixed point iteration \cite{vogel} (see algorithm \ref{FP_Iter}) to solve regularized LS optimization problem with penalties (\ref{wcr.H},\ref{wcr.HTV}) and (\ref{wcr.HTVL2}).

\begin{algorithm}
\caption{Lagged Diffusivity Fixed Point Method for Regularized LS}
\label{FP_Iter}
\begin{algorithmic}[15]
\\ $\nu := 0$
\\ $\bm{u_{0}} := $ initialization 
\\ begin outer (fixed point) iterations
\\ \ \ \ $\bm{g}_{\nu} := \mathrm{A}^{*}(\mathrm{A}\bm{u} - \bm{b}) + \alpha\mathrm{R}\bm{u};$  \ \ \ $\%$ \textit{gradient}
\\ \ \ \ $\mathrm{H} := \mathrm{A}^{*}\mathrm{A} + \alpha\mathrm{R};$  \ \ \ $\%$ \textit{approximate Hessian}
\\ \ \ \ $\bm{s}_{\nu} := -\mathrm{H}^{-1}\bm{g}_{\nu};$  \ \ \ $\%$ \textit{quasi-Newton step}
\\ \ \ \ $\bm{u}_{\nu + 1} := \bm{u}_{\nu} + \bm{s}_{\nu};$  \ \ \ $\%$ \textit{update approximate solution}
\\ \ \ \ $\nu = \nu + 1;$
\\ end outer (fixed point) iterations
\end{algorithmic}
\end{algorithm}
The system $\mathrm{H}\bm{s}_{\nu}  = \bm{g}_{\nu}$  is solved by usual CG method \cite{nocedal}, let $l$ be an iteration index and $L$ is the maximum number of inner iterations for CG method. Inner and outer iterations can be terminated earlier if the following stopping criteria met $\left\Vert\bm{s}_{l} - \bm{s}_{l + 1} \right\Vert_{2}^{2} \leq \rho$ and $\left\Vert\bm{u}_{\nu} - \bm{u}_{\nu + 1} \right\Vert_{2}^{2} \leq \rho$ respectively. We chose $\rho=10^{-4}$ to be a small constant.

The largest eigenvalue of the matrix $\mathrm{R}$ is of the order ${\cal O}(h^{-2})$, $h = \min(h_x, h_y)$, and the smallest are of the order ${\cal O}(1)$. If $\alpha$ is not very small (considerably larger than $h^4$), then the large condition number of $\mathrm{R}^{*}\mathrm{R}$ can slow down the convergence of CG iterations for the minimization of $\psi_\alpha(\bm{u})$. To alleviate this problem, we introduce preconditioning that consists in the multiplication of the gradient of $\psi_\alpha(\bm{u})$ by the inverse of $\mathrm{H}= \sigma^2 \mathrm{I} + \alpha \mathrm{R}$ in the course of CG iterations, where $\sigma$ is a scalar value of the order of the largest singular value of $\mathrm{A}$. Since matrix $\mathrm{H}_\sigma$ is very sparse, the application of its inverse to a vector can be efficiently performed (via the factorization of $\mathrm{H}_\sigma$) by modern state-of-the art sparse direct solvers \cite{davis}.

\subsection{Regularized reconstruction with Poisson data fidelity term}
To solve regularized Poisson log-likelihood problem (\ref{kl.reg}) we use the splitting technique introduced in \cite{sawatsky} and used for SPECT reconstruction in \cite{kazantsev}. The main idea is to decouple data-fidelity term from the regularizer by solving two problems independently (see algorithm \ref{Split_Rec}).
\begin{algorithm}
\caption{Splitting algorithm for regularized Poisson term}
\label{Split_Rec}
\begin{algorithmic}[15]
\\ $\nu := 0$
\\ $\bm{u_{0}} := $ initialization with ones
\\ begin outer iterations
\\ \ \ \ $\bm{u}_{\nu + \frac{1}{2}} := \bm{u}_{\nu}./(\mathrm{A}^{*} \bm{1}).*\mathrm{A}^{*} (\bm{b}./\mathrm{A}\bm{u}_{\nu});$  \ \ \ $\%$ \textit{MLEM step}
\\ \ \ \ \ \ $\bm{f_{0}} := \bm{u}_{\nu + \frac{1}{2}};$  \ \ \ 
\\ \ \ \ \ \ $l := 0;$  \ \ \ 
\\ \ \ \ \ \ $\bm{f_{l+1}} := \bm{f_{l}} + \tau((\bm{f_{l}} - \bm{f_{0}}) + \mathrm{R} \bm{f_{l}});$  \ \ \ $\%$ \textit{image denoising}
\\ \ \ \ $\bm{u}_{\nu + 1} := \bm{f_{l+1}};$  \ \ \ $\%$ \textit{update approximate solution}
\\ \ \ \ $\nu = \nu + 1;$
\\ end outer iterations
\end{algorithmic}
\end{algorithm}
From the structure of algorithm \ref{Split_Rec}, one can see that the optimization problem with Poisson log-likelihood term is solved independently with Maximum Likelihood Expectation Maximization (MLEM) algorithm \cite{vogel}. Regularization is performed as an additional image denoising step. In MLEM step, $\bm{1}$ denotes the vector of 1's, and $.*$ and $./$ denote componentwise multiplication and division, respectively. Additionally the weights $\bm{w_x}$ and $\bm{w_y}$ for the proposed term (\ref{wcr.H}) are calculated after each MLEM step and kept fixed in image denoising updates.  
 

\section{Numerical Results}\label{sec:results}
In this section we present two different numerical experiments with the proposed  penalty EL (\ref{wcr.H}) and compare it with TV (\ref{wcr.HTV}) and TV-$\ell_{2}$  (\ref{wcr.HTVL2}) penalties. In the first experiment we model reconstruction of the synthetic phantom with algorithm \ref{FP_Iter} and in the second experiment we perform reconstruction of more realistic phantom for emission tomography with algorithm \ref{Split_Rec}. The MATLAB code for tomographic reconstruction using TV  (\ref{wcr.HTV}) regularization and the proposed EL penalty (\ref{wcr.H}) is provided \cite{code}.

\subsection{CT reconstruction}
\begin{figure}[ht]
  \centering
  {\includegraphics[width=0.3\textwidth]{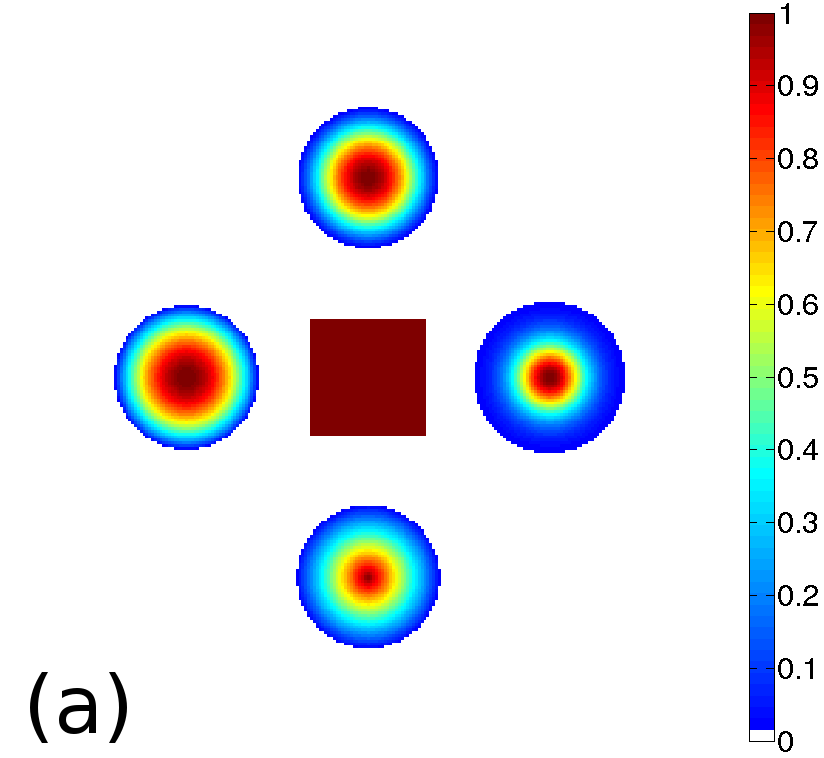}}
  {\includegraphics[width=0.4\textwidth]{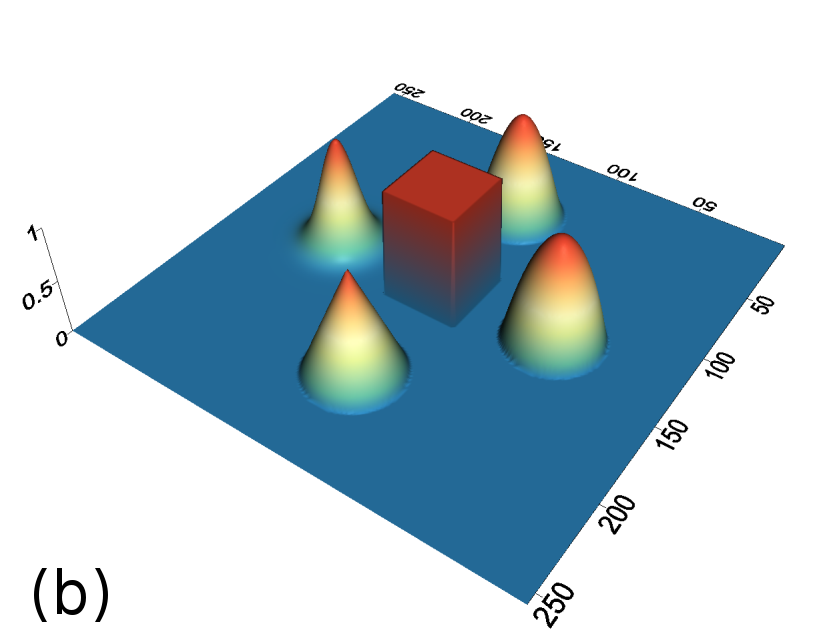}} 
  \caption{(a) 2D analytical phantom which consists of 2 parabolas, 2 Gaussians and a rectangle; (b) a surface representation of the phantom.}
  \label{phantom}
\end{figure}

To test the proposed penalty in regularized tomographic reconstruction we designed an analytical phantom which consists of a smooth (two Gaussians and two parabolas) and piecewise-constant (one rectangle) functions (see Fig. \ref{phantom}). 

To avoid of reconstructing on the same grid where projection data was generated (so-called reconstruction with ``inverse crime'' \cite{hansen}), we used a higher resolution of the phantom on a $500 \times 500$ isotropic pixel grid to generate projections with a strip kernel \cite{kak}. Then Poisson distributed noise was applied to the projection data, assuming an incoming beam intensity of 3$\cdot 10^{5}$ (photon count).  Reconstructions were calculated on a $250 \times 250$ isotropic pixel grid and with a linear projection model \cite{kak}. We used 90 projection angles in 180 degrees (assuming a parallel beam geometry) to reconstruct the phantom. 

\begin{figure}[ht]
  \centering
{\includegraphics[width=0.32\textwidth]{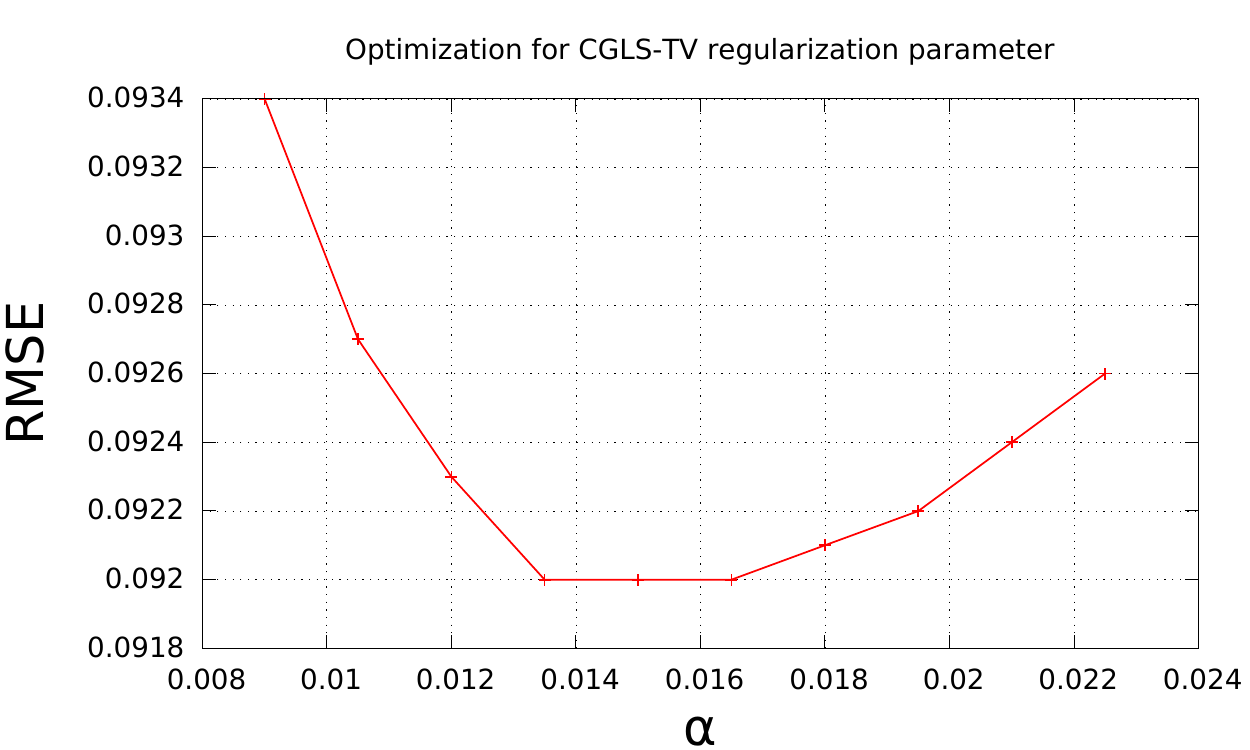}}
{\includegraphics[width=0.32\textwidth]{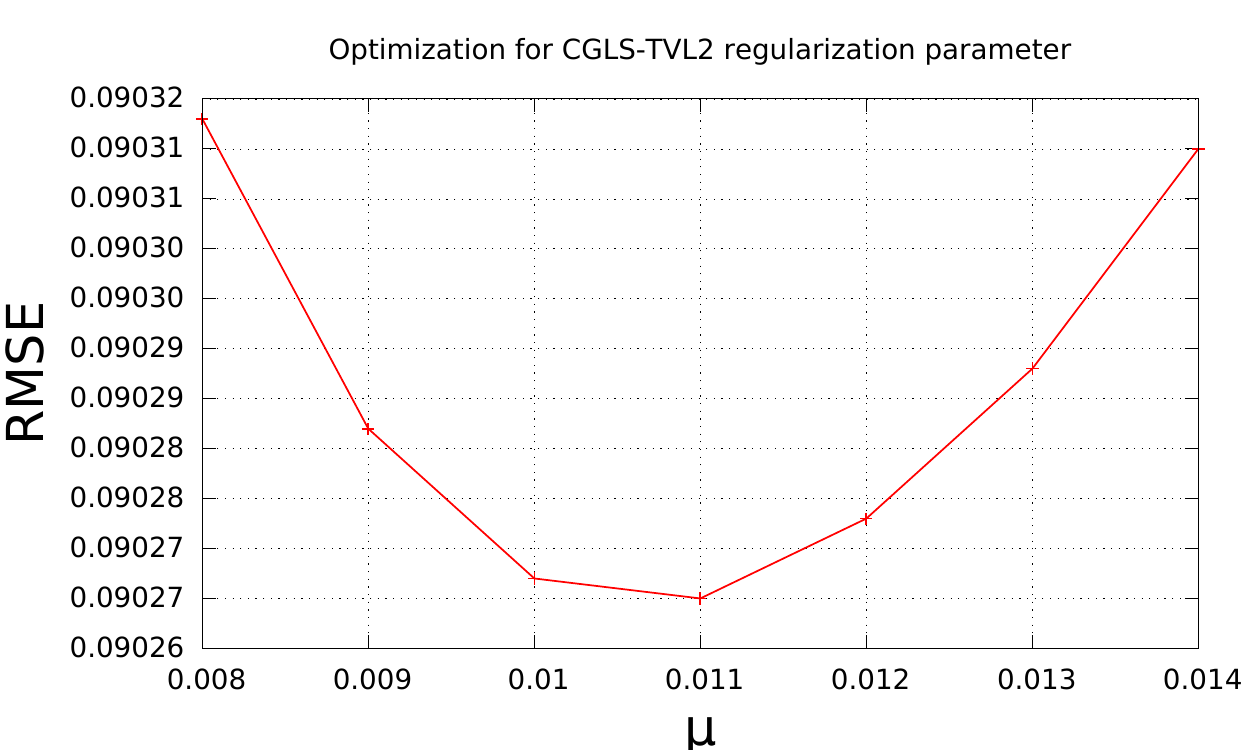}}          {\includegraphics[width=0.32\textwidth]{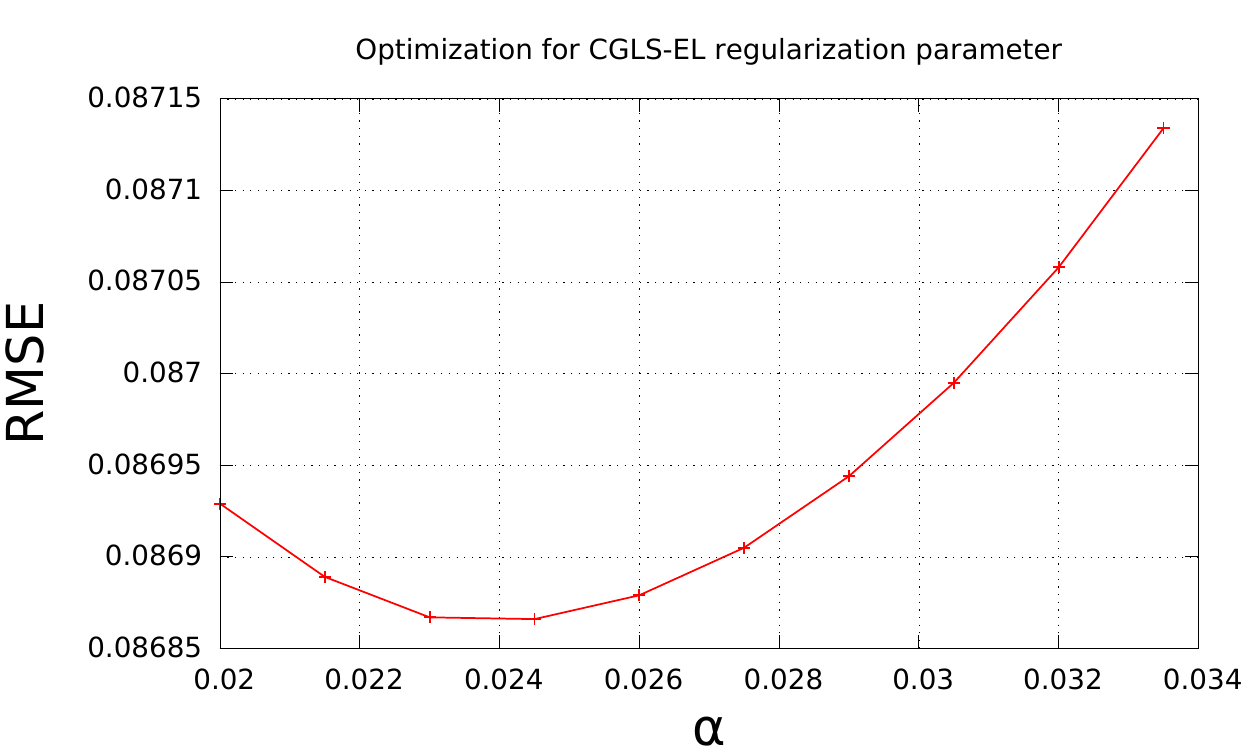}} 
  \caption{Optimization procedure to find the optimal regularization parameters for TV (left), TV-$\ell_{2}$ (middle) and EL (right) algorithms. For  TV-$\ell_{2}$ penalty we optimized the second regularization constant ($\mu$) while keeping the optimal $\alpha$ fixed (found for TV previously).}
  \label{reg_param}
\end{figure}

For a fair comparison of different regularizing penalties we initially optimized the regularization parameters (see Fig. \ref{reg_param}) with respect to the value of root-mean-square-error (RMSE), defined as:
\begin{equation}
\textrm{RMSE}(\overline{\bm{u}}, \hat{\bm{u}}) =  \frac{\left\Vert \hat{\bm{u}} - \overline{\bm{u}} \right\Vert_{2}}{\left\Vert \overline{\bm{u}} \right\Vert_{2}},
\label{MSE}
\end{equation}
where $\overline{\bm{u}}$ is an exact image and $\hat{\bm{u}}$ is a reconstructed image.

\begin{figure}[ht]
  \centering
{\includegraphics[width=0.6\textwidth]{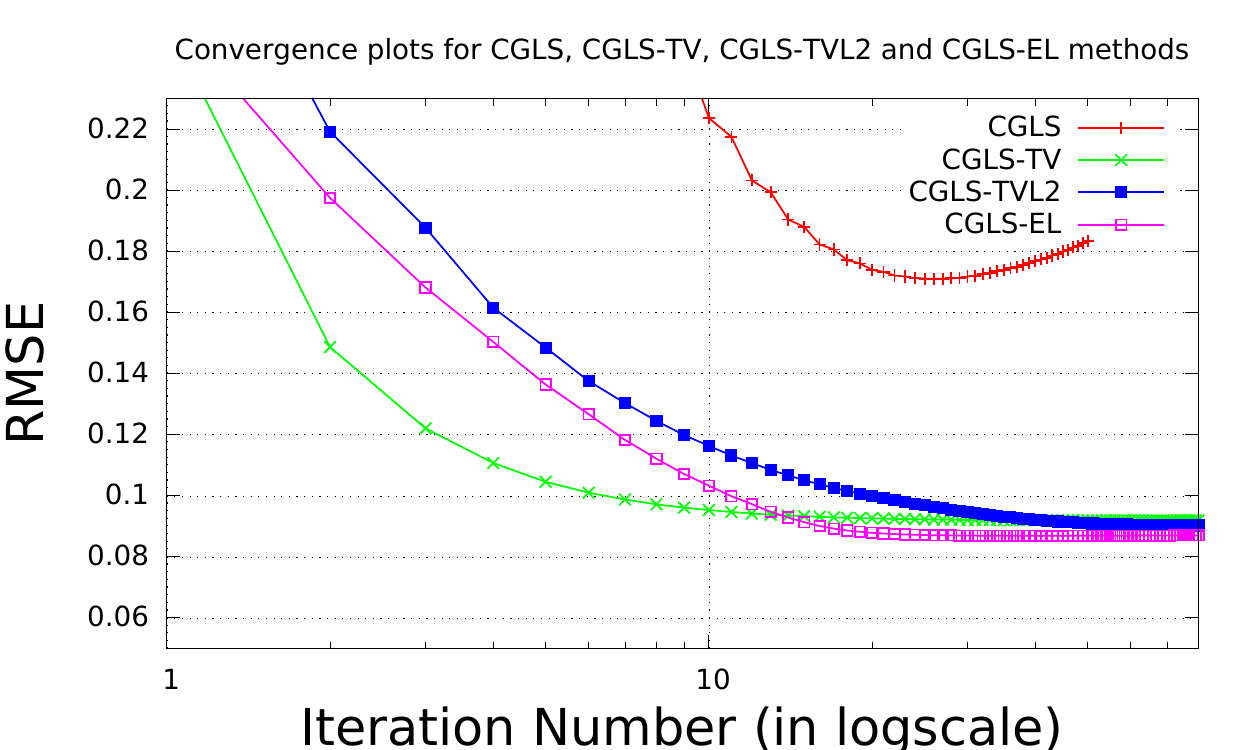}}
  \caption{Plot of RMSE values to the number of iterations for CGLS, CGLS-TV, CGLS-TV-$\ell_{2}$ and CGLS-EL methods. CGLS-EL method outperforms other algorithms (see table \ref{tab:RMSE1}).}
  \label{iter_alg}
\end{figure}

We found empirically that $\beta = 0.03$ for EL penalty (\ref{wcr.H}) gives good results for the presented experiments, therefore we will keep it fixed for the rest of our tests. With fixed optimal regularization parameters (see Fig. \ref{reg_param}) we perform $L = 80$ outer (fixed point) iterations and 5 inner iterations of algorithm \ref{FP_Iter} with different penalties (see Fig. \ref{iter_alg}).
In Fig. \ref{iter_alg} one can see that the CGLS method diverges quickly  while regularized methods converge to a steady-state solution. The lowest RMSE value is achieved with the proposed EL penalty (see table \ref{tab:RMSE1}). Reconstruction with TV penalty gives the highest RMSE value (from all compared regularized methods) however it shows faster convergence on first fixed point iterations. TV-$\ell_{2}$ penalty performs slightly better than TV, but still with higher RMSE than EL method. 

\begin{table}[ht] 
	\caption{RMSE values for CGLS, CGLS-TV, CGLS-TV-$\ell_{2}$ and CGLS-EL methods} 	
	\centering
    \begin{tabular}{| l | l | l | l | l |}
	\hline
	  & CGLS & CGLS-TV & CGLS-TV-$\ell_{2}$ & CGLS-EL  \\ \hline
    RMSE & 0.1710 & 0.0919 & 0.0902 & 0.0868 \\
    \hline
    \end{tabular}
    \label{tab:RMSE1} 
\end{table}

\begin{figure}[ht]
  \centering
{\includegraphics[width=0.23\textwidth]{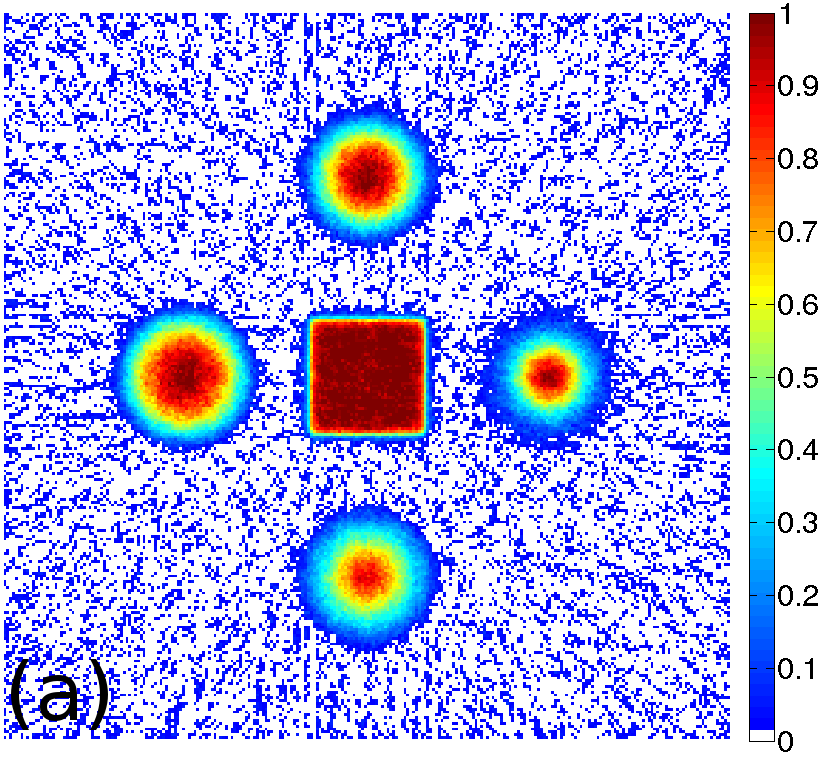}}
{\includegraphics[width=0.23\textwidth]{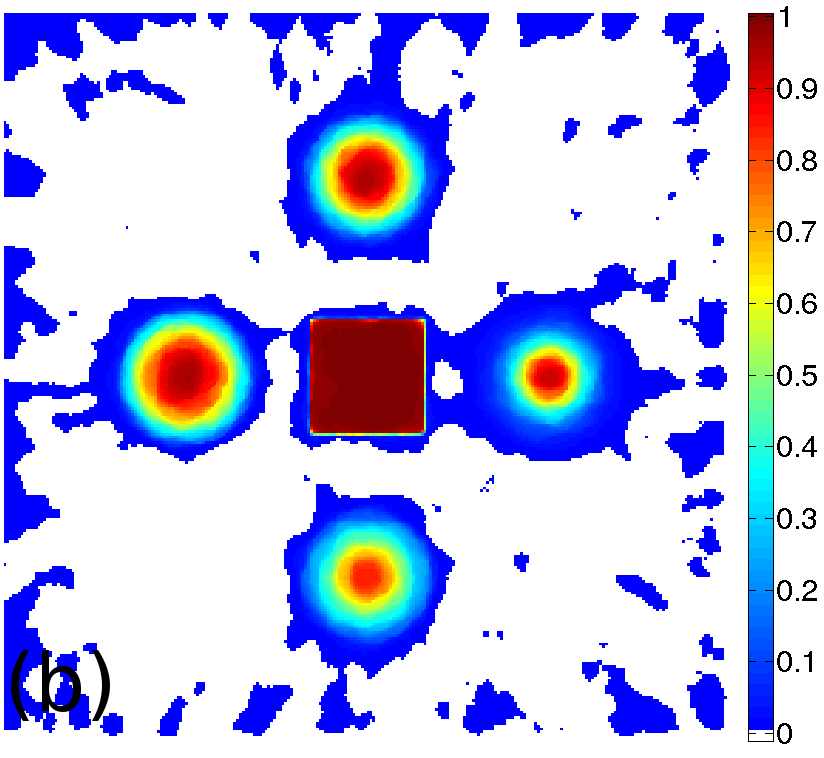}}          {\includegraphics[width=0.23\textwidth]{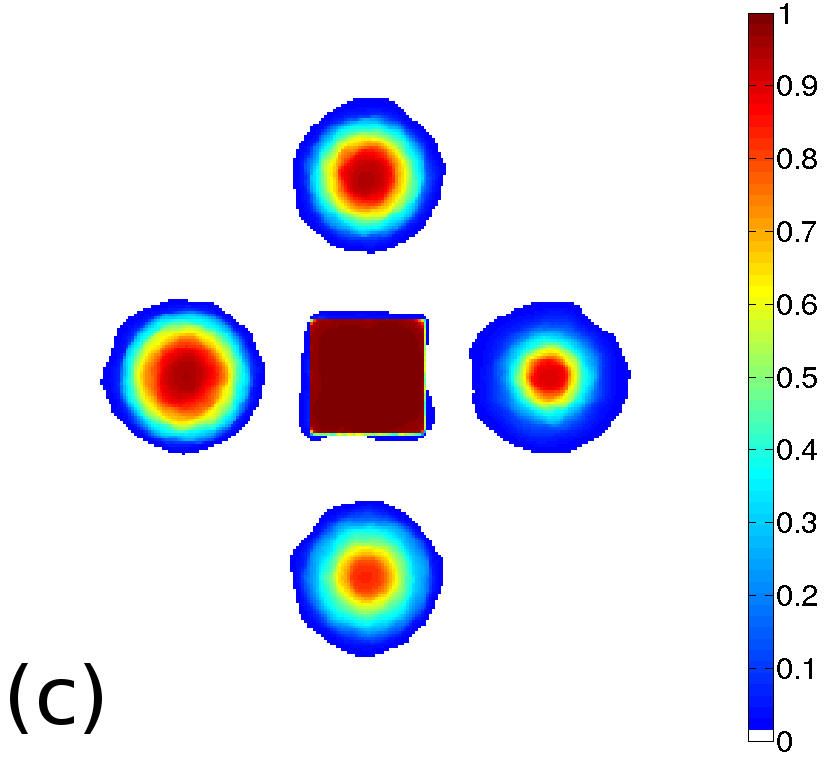}} 
{\includegraphics[width=0.23\textwidth]{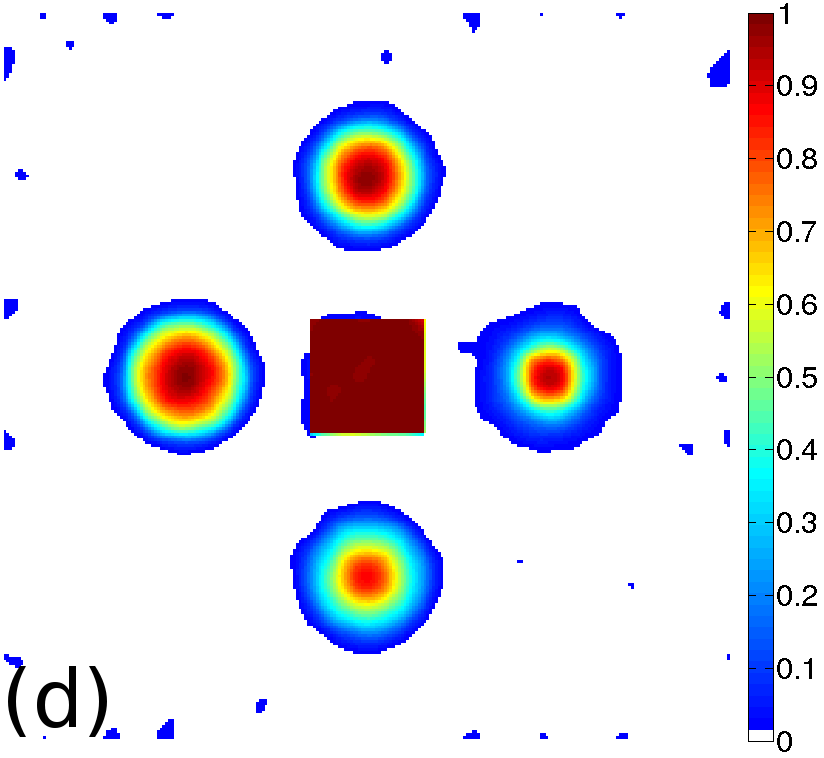}} 
  \caption{Reconstructed phantoms with (a) CGLS, (b) CGLS-TV, (c) CGLS-TV-$\ell_{2}$ and (d) CGLS-EL method.}
  \label{rec1}
\end{figure}

\begin{figure}[ht]
  \centering
{\includegraphics[width=0.24\textwidth]{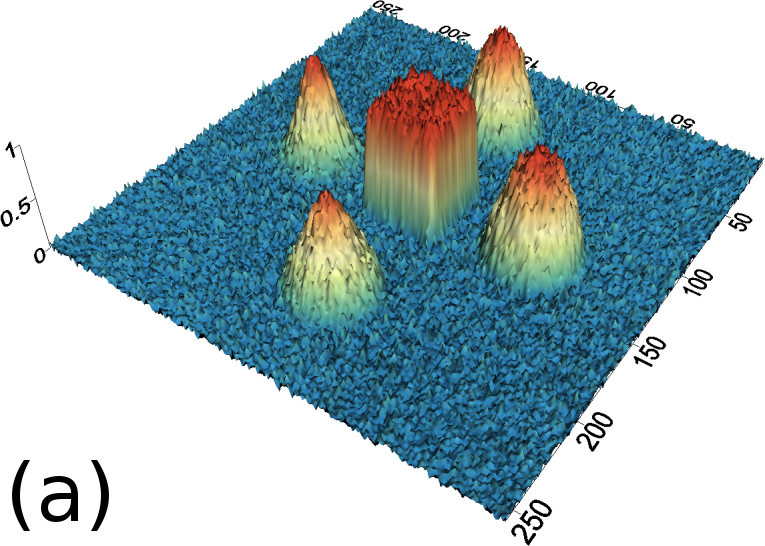}}
{\includegraphics[width=0.24\textwidth]{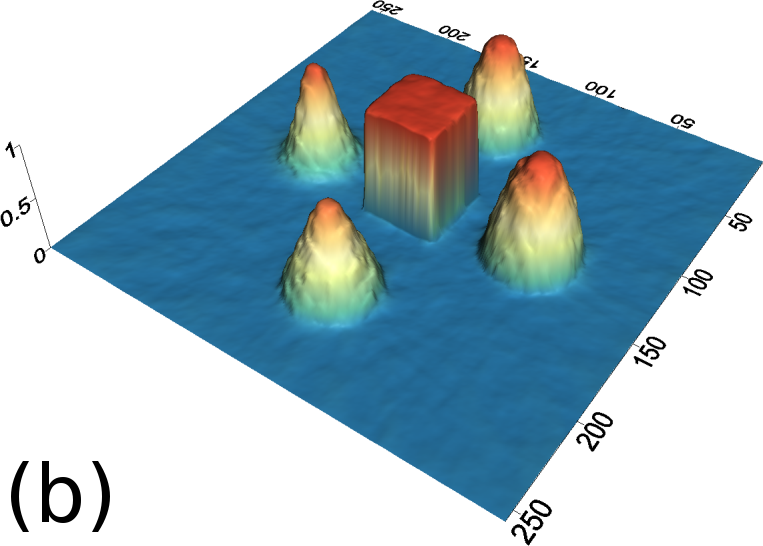}}          {\includegraphics[width=0.24\textwidth]{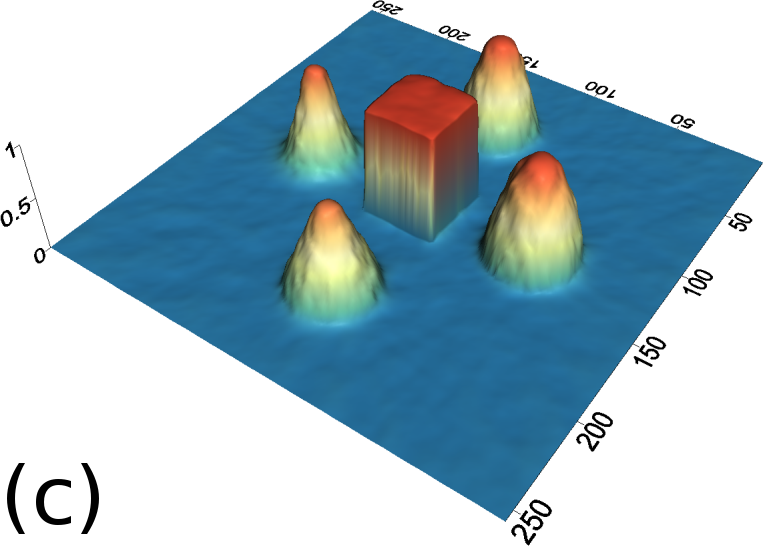}} 
{\includegraphics[width=0.24\textwidth]{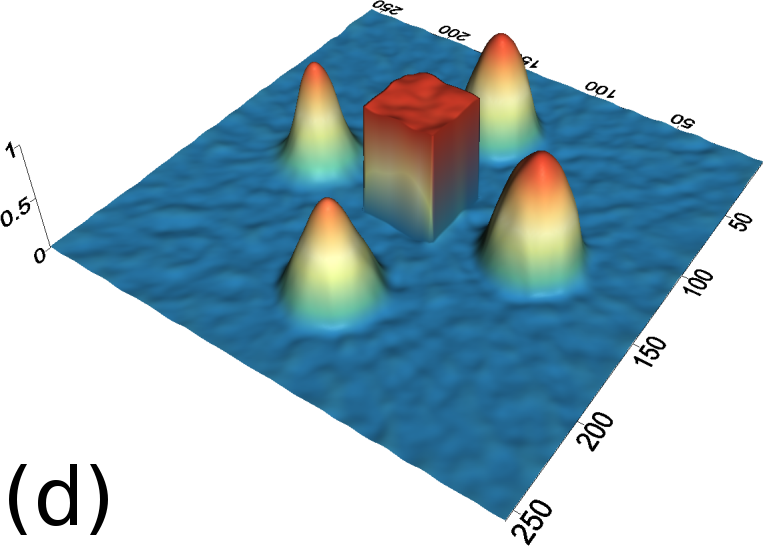}} 
  \caption{Surface representations of the reconstructed phantoms with (a) CGLS, (b) CGLS-TV, (c) CGLS-TV-$\ell_{2}$ and (d) CGLS-EL method.}
  \label{rec2}
\end{figure}

\begin{figure}[ht]
  \centering
{\includegraphics[width=0.3\textwidth]{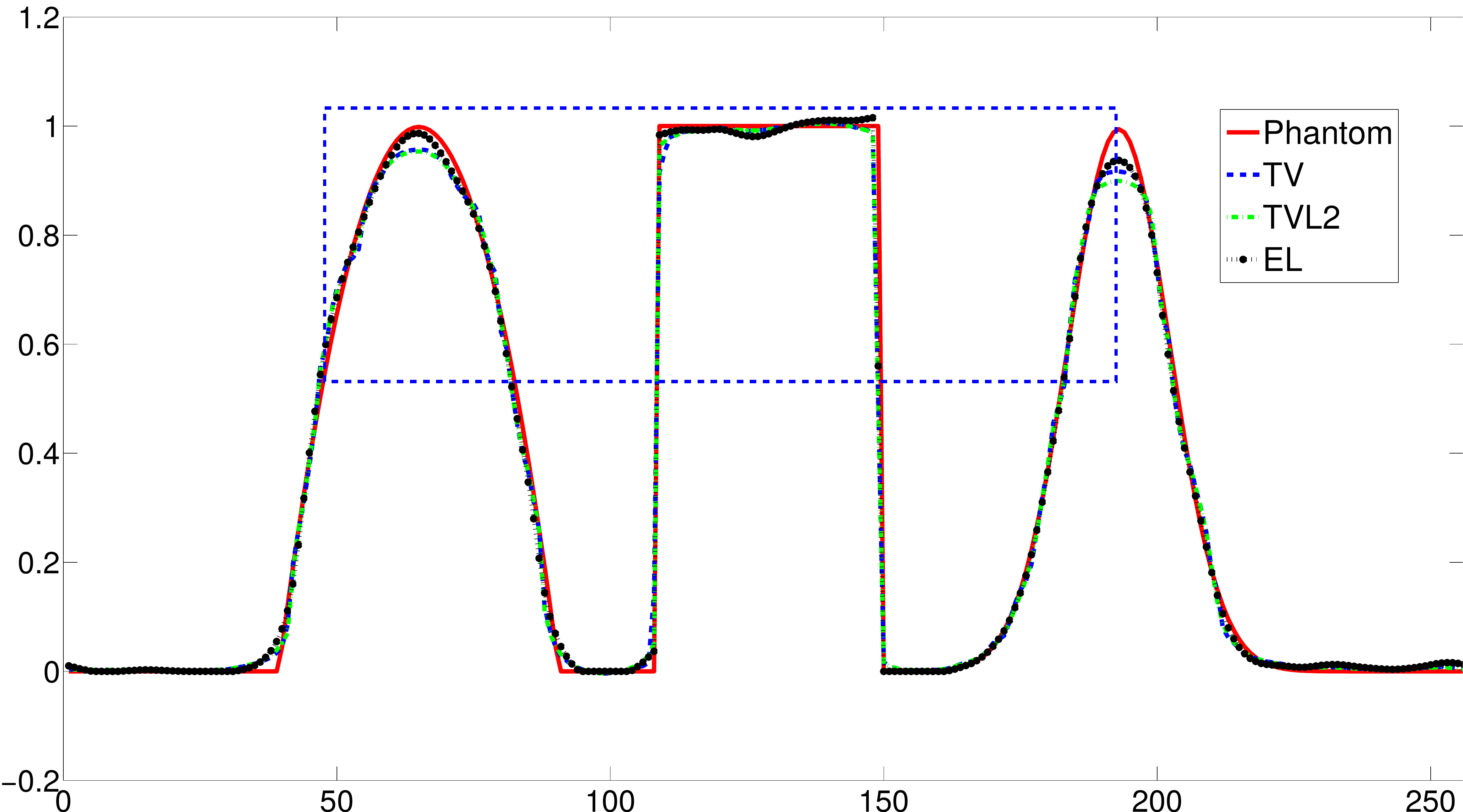}}
{\includegraphics[width=0.4\textwidth]{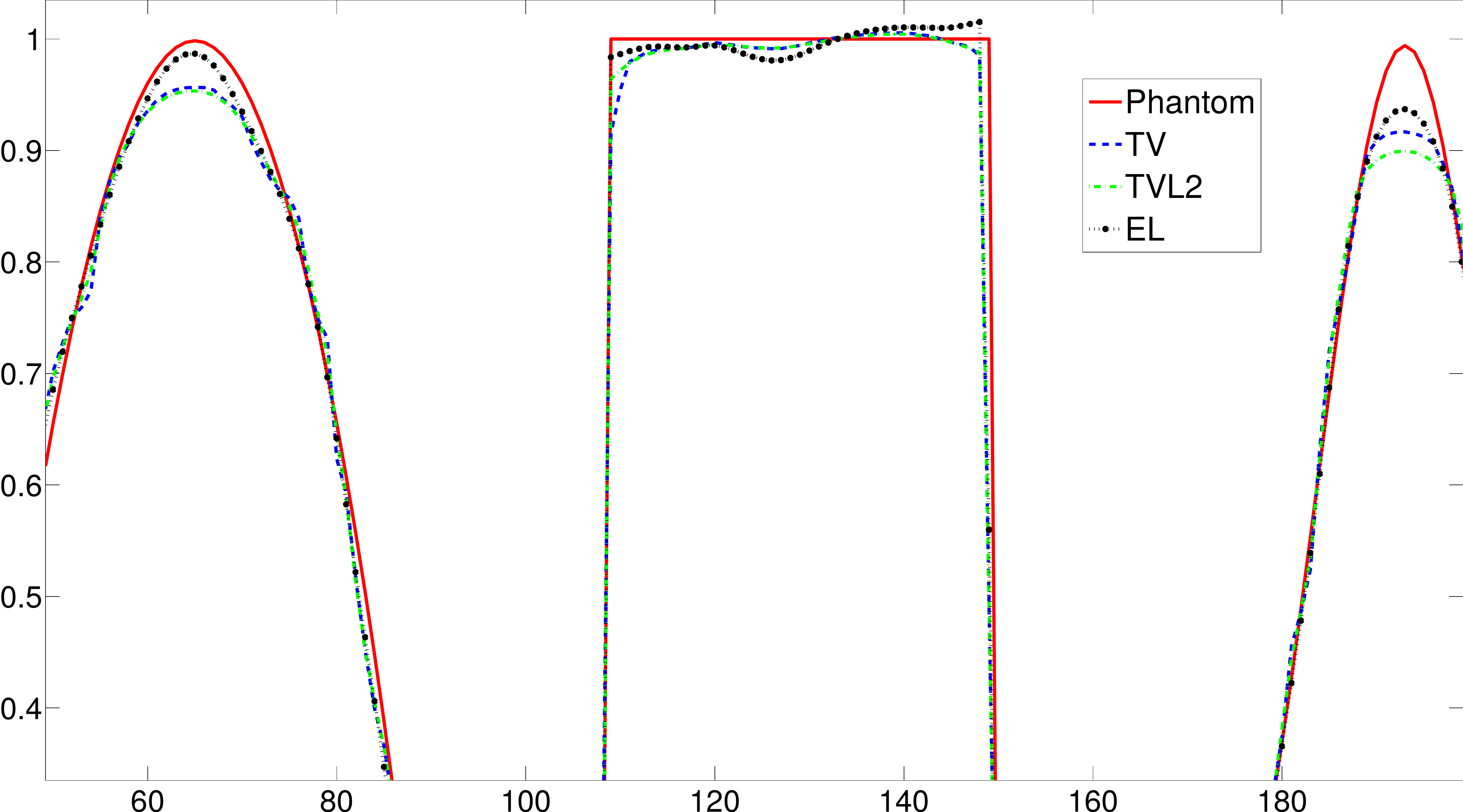}} \\         {\includegraphics[width=0.3\textwidth]{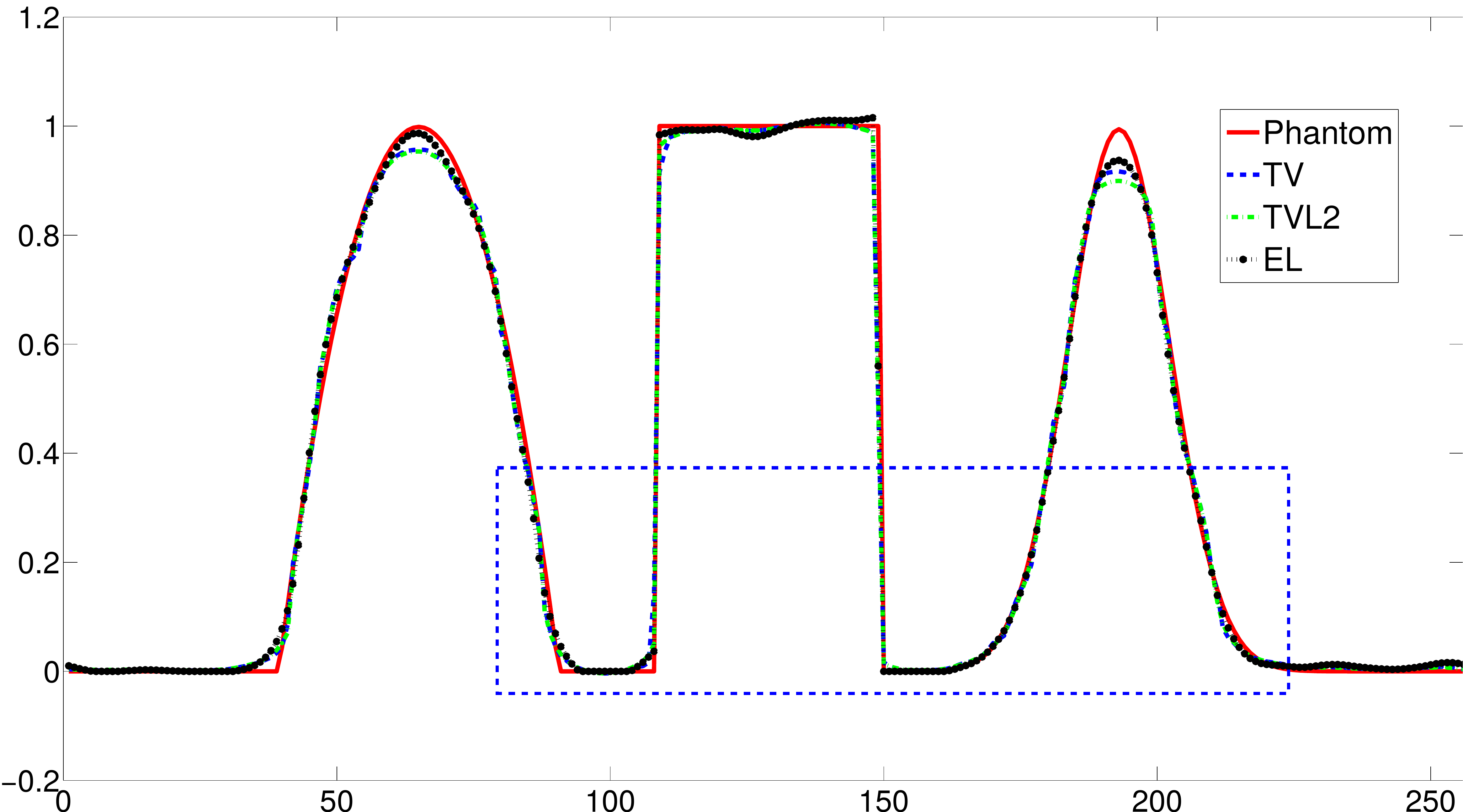}} 
{\includegraphics[width=0.4\textwidth]{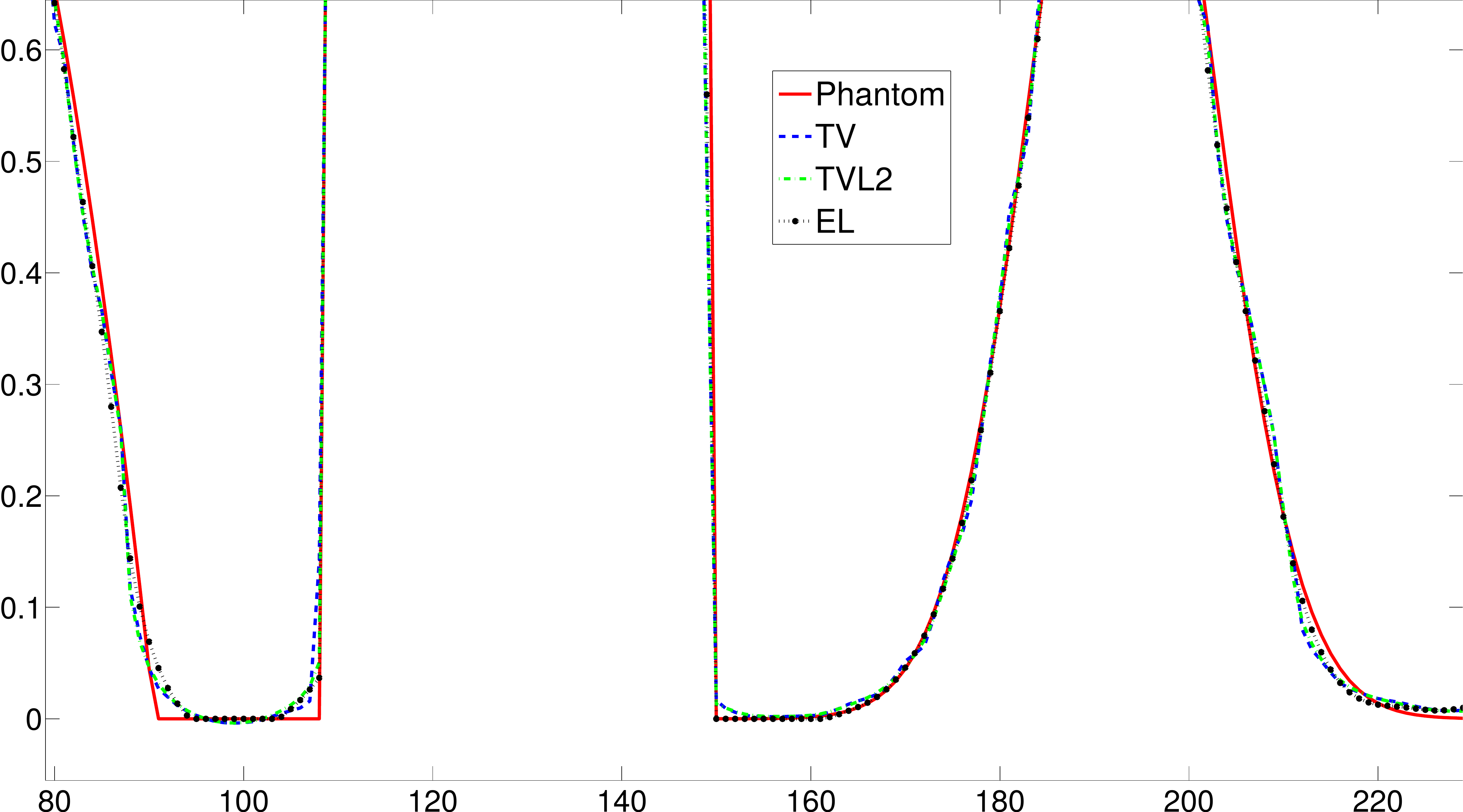}} 
  \caption{Horizontal middle cross-sections of the phantom and reconstructed images (see Fig. \ref{rec1}). Left side images show the region selected and the right side images show the magnified region. It can be clearly seen that the EL penalty performs much better for smooth objects while slightly more perturbed in uniform areas (e.g. the top of the rectangle).}
  \label{rec2_plot}
\end{figure}

Reconstructed images are presented in Fig. \ref{rec1}. Since CGLS-TV-$\ell_{2}$ reconstruction might look more appealing than CGLS-EL we also show the surface  representations of reconstructed images (see Fig. \ref{rec2}) and horizontal middle cross-sections (see Fig. \ref{rec2_plot}).

One can notice that CGLS reconstruction is very noisy. CGLS-TV method better suppresses noise, however smooth features are strongly affected by the ``staircasing'' effect. CGLS-TV-$\ell_{2}$ method provides reconstruction with smoother features and CGLS-EL method resolves smooth features even better (e.g. cone-shaped parabola). Although CGLS-EL method performs very well for smooth objects one can notice the wave-like variations of intensity in the background and also at the top of the rectangle (see Fig. \ref{rec2_plot}). This issue can be explained by the properties of our regularizer, in contrast to TV, our penalty does not seek the sparsest solution and does not penalize strongly (pushing to the constant value) a small intensity perturbations. The EL term tends to preserve all sharp edges while uniform noise is smoothed isotropically with the Laplacian. In Fig. \ref{rec2_plot} one can see that the CGLS-EL method provides better recovery of smooth features while slightly higher (compare to TV and TV-$\ell_{2}$) perturbations visible in uniform areas (the top of the rectangle), however, the edges of the rectangle are defined sharper with the EL penalty. 

\subsection{ET reconstruction}
\begin{figure}[ht]
  \centering
{\includegraphics[width=0.25\textwidth]{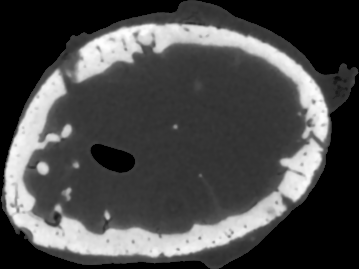}}
{\includegraphics[width=0.25\textwidth]{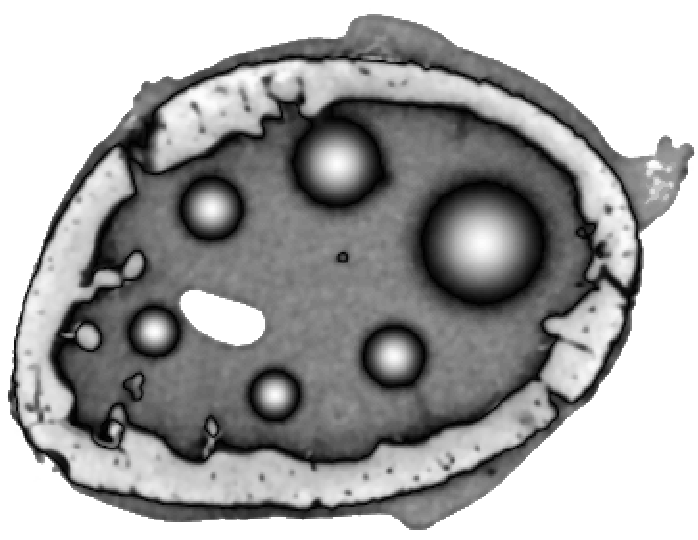}}
{\includegraphics[width=0.25\textwidth]{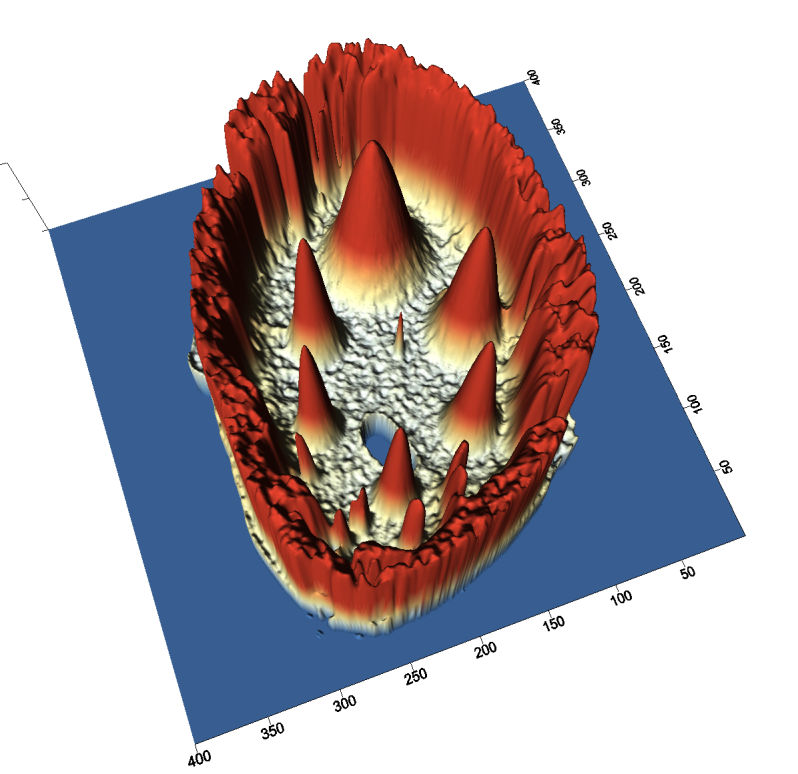}}       
  \caption{(left) High quality X-ray reconstruction of a mice bone; (middle) phantom created from the reconstruction (left) with a six Gaussians added to it; (right) a surface representation of the phantom.}
  \label{ET_phant}
\end{figure}

To simulate emission tomography reconstruction we designed a more realistic phantom from the high-quality X-ray scan of a mice bone. The data was acquired on a Nikon Metris Custom Bay cone-beam scanner at the Henry Moseley Manchester X-ray facility, and was reconstructed with the Feldkamp algorithm (see Fig. \ref{ET_phant} (left)). We thresholded the obtained reconstruction and added six gaussians with various kernel widths (see Fig. \ref{ET_phant} (middle and right)). 

To simulate PET projection data we used NiftyRec \cite{pedemonte}, a software for tomographic reconstruction, providing GPU-accelerated reconstruction for emission and transmission computed tomography. The phantom size is $400 \times 400$ pixels and 300 projections was simulated. Poisson noise was added to projections with an expected number of $10\cdot 10^{6}$ photon counts in total. Twenty noise realizations were simulated to estimate methods quantitatively. The point spread function of the PET system was modelled (with convolution of the sinogram columns with a Gaussian of full width half maximum of three pixels) in the projection and back-projection operations. No scatter was simulated in this study. For our experiments (see algorithm \ref{Split_Rec}) we performed 130 MLEM iterations and 5 inner iterations (denoising step).

\begin{figure}[ht]
  \centering
{\includegraphics[width=0.32\textwidth]{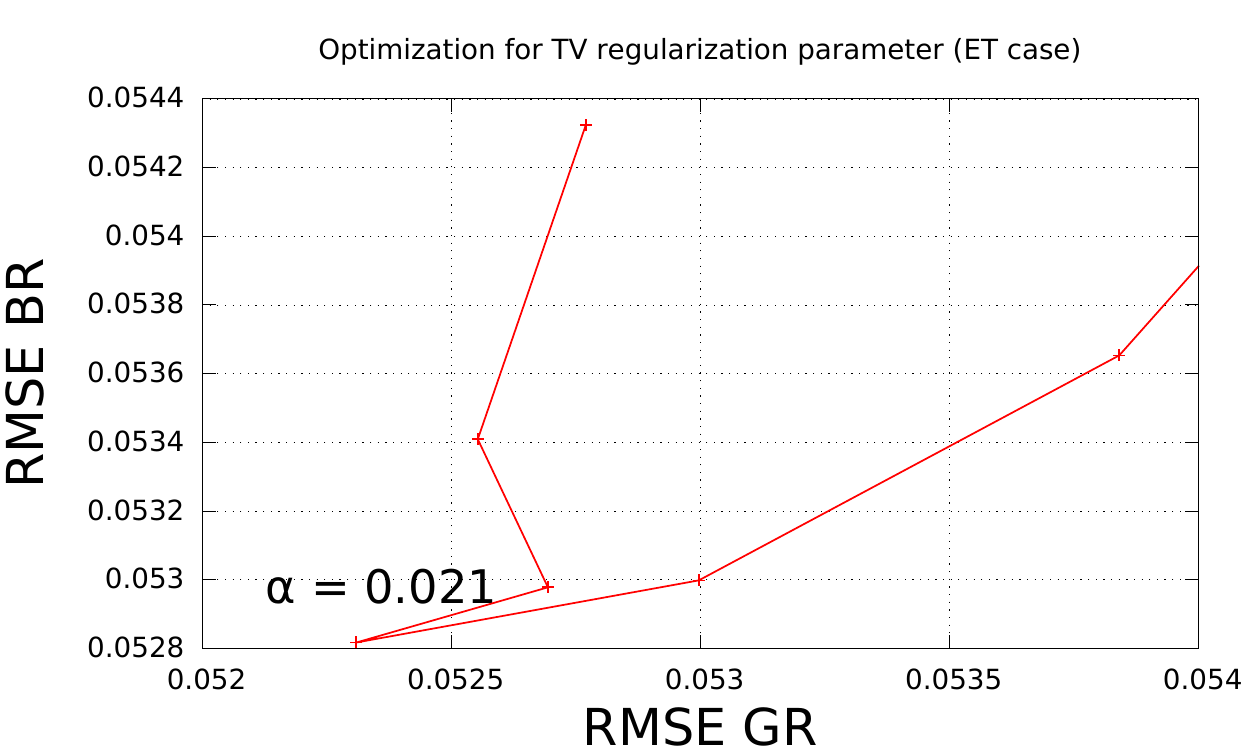}}
{\includegraphics[width=0.32\textwidth]{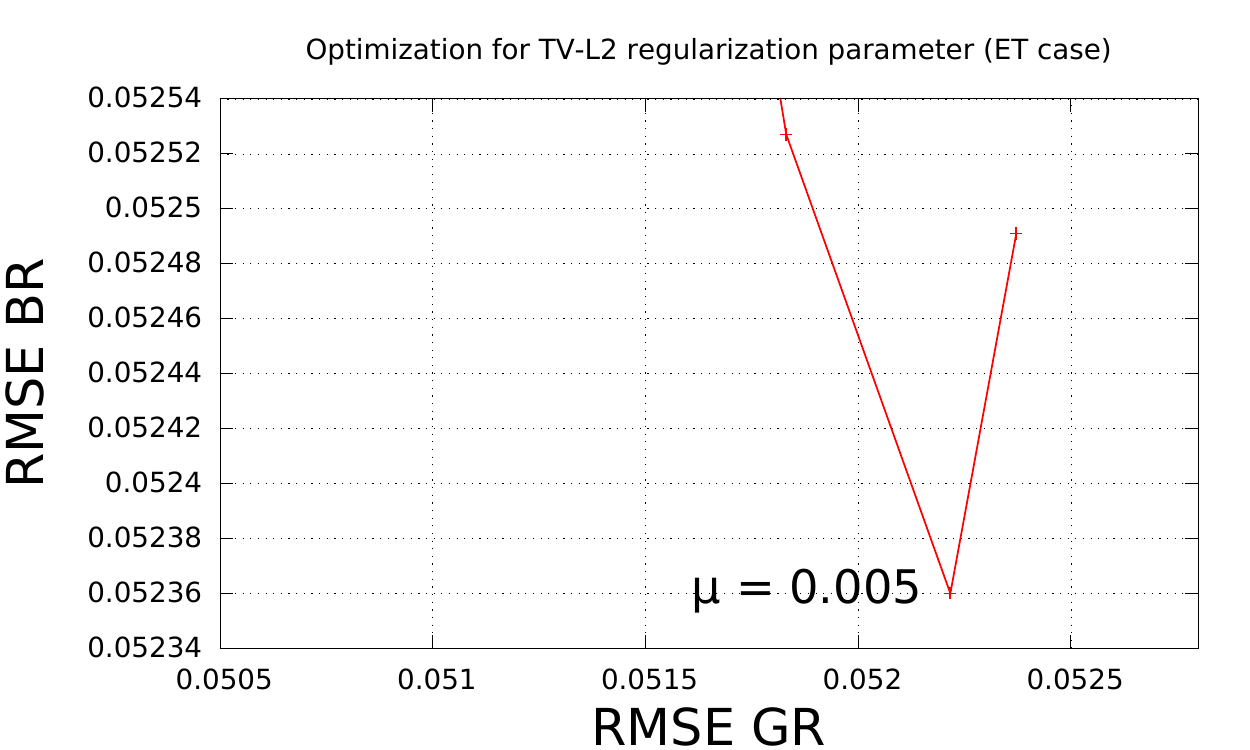}}
{\includegraphics[width=0.32\textwidth]{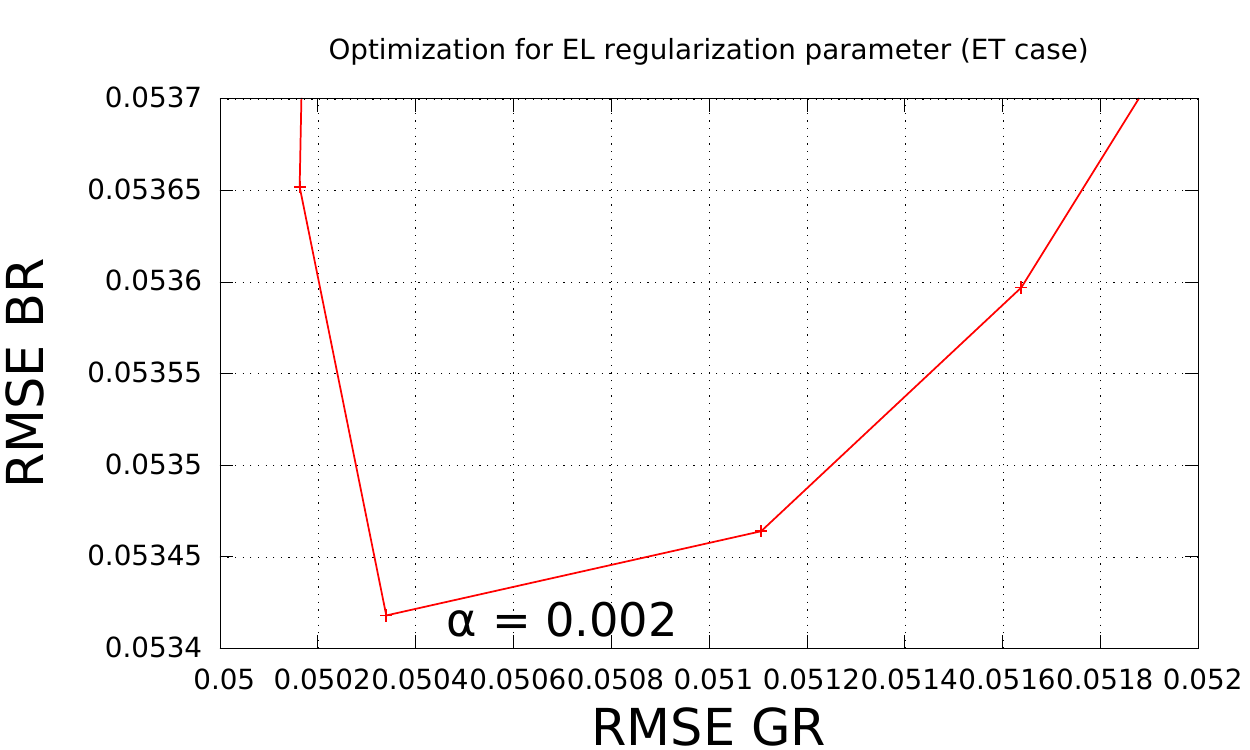}}       
  \caption{Optimization procedure to find the optimal regularization parameters for MLEM-TV (left), MLEM-TV-$\ell_{2}$ (middle) and MLEM-EL (right) algorithms. For  MLEM-TV-$\ell_{2}$ method we optimized the second regularization constant ($\mu$) while kept the optimal $\alpha$ fixed (found for MLEM-TV method previously).}
  \label{ET_regul}
\end{figure}

To quantify obtained reconstructions we used averaged over all noise realizations RMSE (\ref{MSE}) values in the bone region (BR) and in Gaussian regions (GR). All regularization parameters were carefully selected by comparing the mean of all RMSE values over all noise realizations in GR and BR (see Fig. \ref{ET_regul}). 

\begin{figure}[ht]
  \centering
{\includegraphics[width=0.38\textwidth]{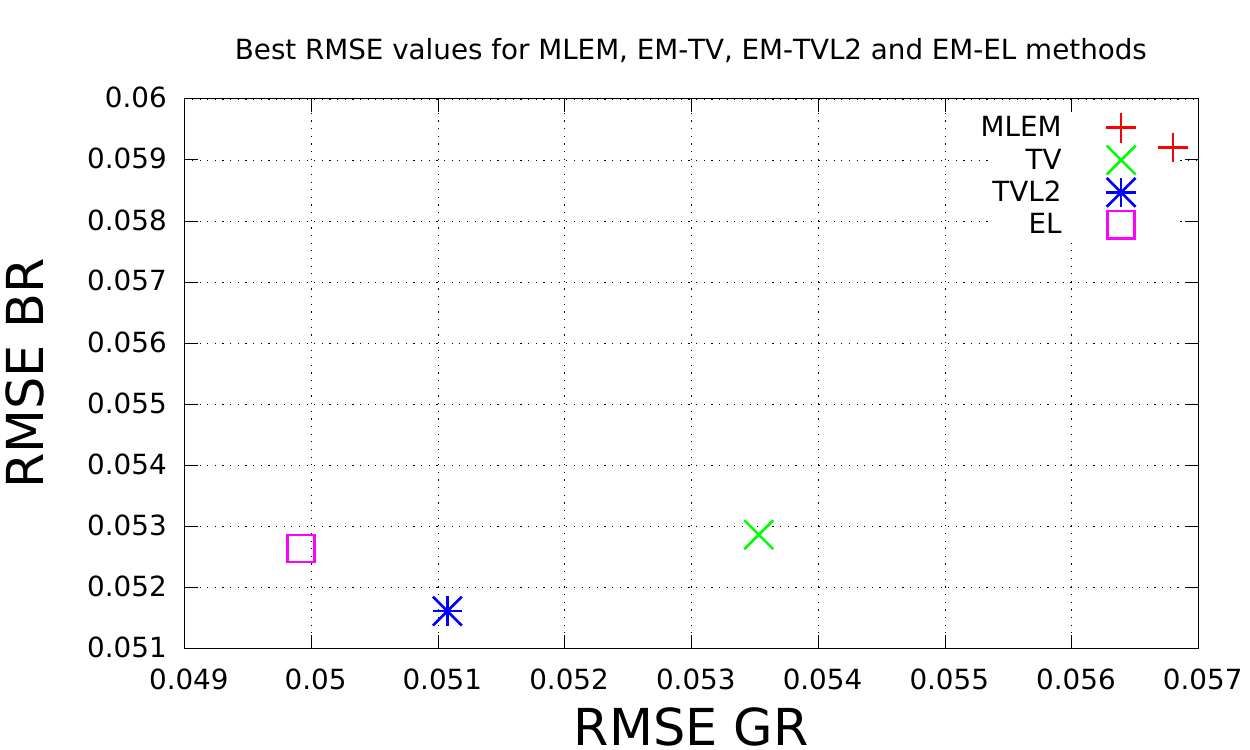}}
  \caption{Plot of  the best RMSE values for MLEM, MLEM-TV, MLEM-TV-$\ell_{2}$ and MLEM-EL methods. MLEM-EL method outperforms other algorithms for GR and gives better result than TV for BR.}
  \label{ET_regul2}
\end{figure}

After estimation of regularization parameters we performed twenty reconstructions for each method with various Poisson noise distributions. The mean values for GR and BR over all noise realizations are shown in Fig. \ref{ET_regul2}. This result proves that the EL penalty is very successful in resolving smooth features (six Gaussians in this case) and also quite competitive for the BR (lower RMSE value than for TV). 

\begin{figure}[ht]
  \centering
{\includegraphics[width=0.23\textwidth]{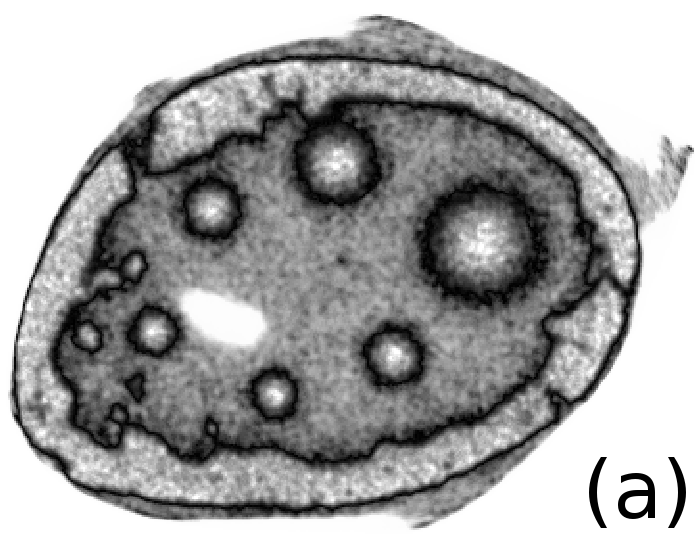}}
{\includegraphics[width=0.23\textwidth]{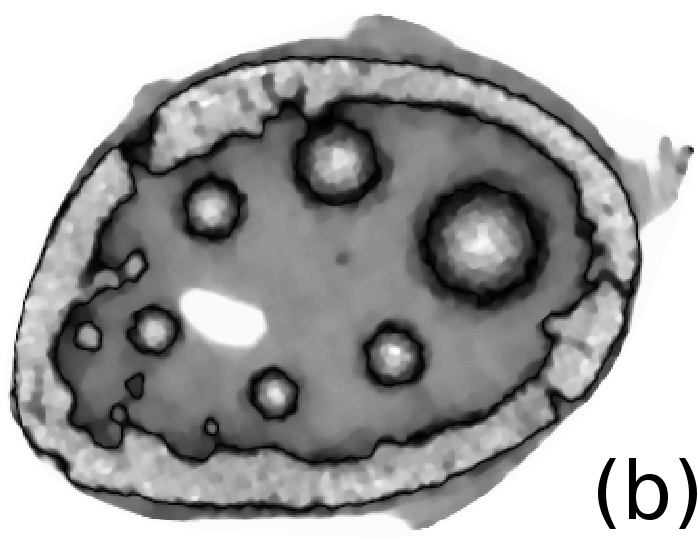}}          {\includegraphics[width=0.23\textwidth]{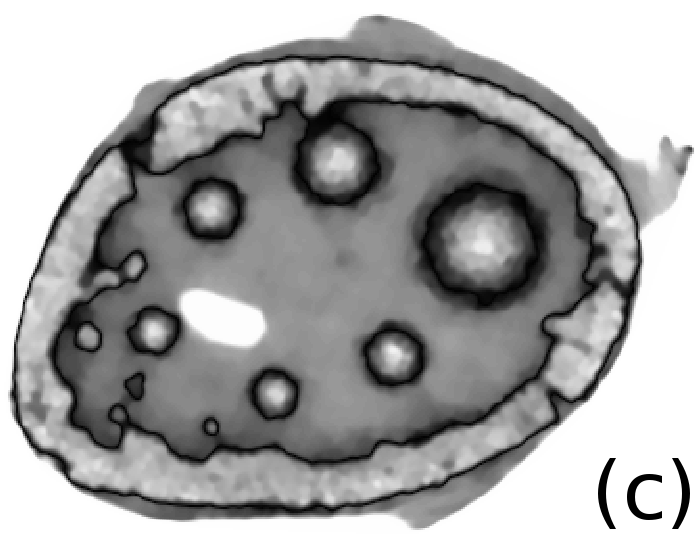}} 
{\includegraphics[width=0.23\textwidth]{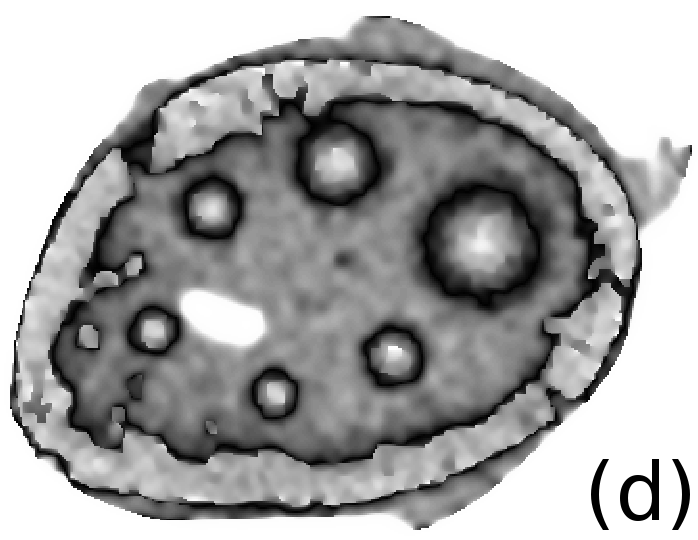}} 
  \caption{Reconstructed phantoms with (a) MLEM, (b) MLEM-TV, (c) MLEM-TV-$\ell_{2}$ and (d) MLEM-EL method.}
  \label{ET_rec1}
\end{figure}

\begin{figure}[ht]
  \centering
{\includegraphics[width=0.24\textwidth]{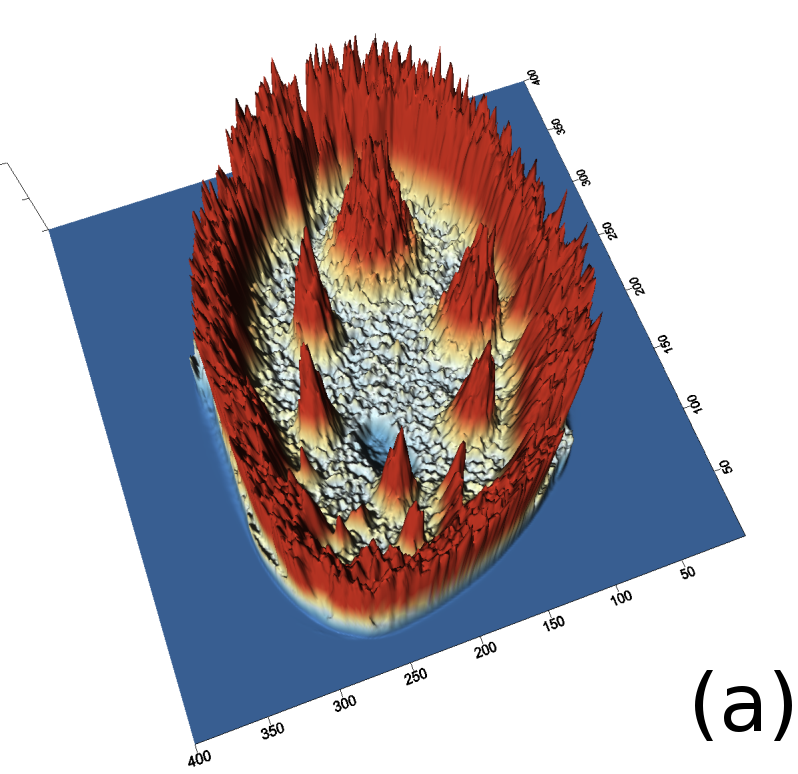}}
{\includegraphics[width=0.24\textwidth]{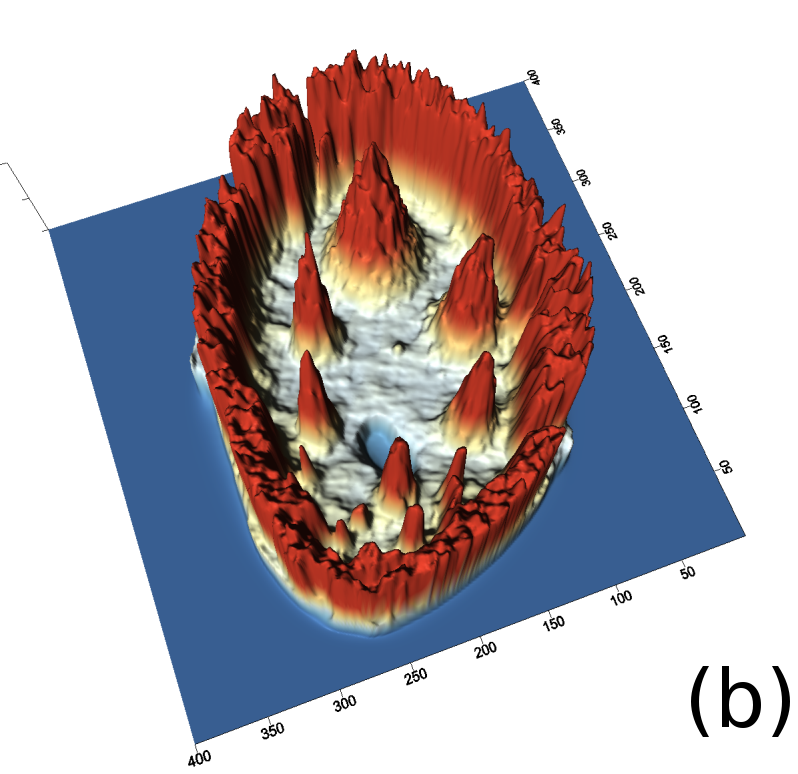}}          {\includegraphics[width=0.24\textwidth]{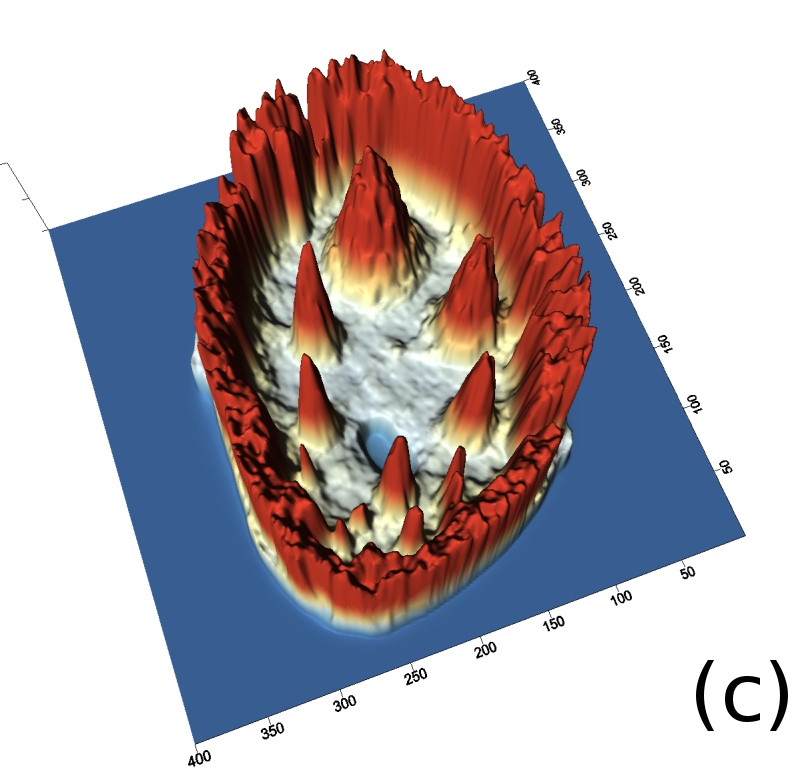}} 
{\includegraphics[width=0.24\textwidth]{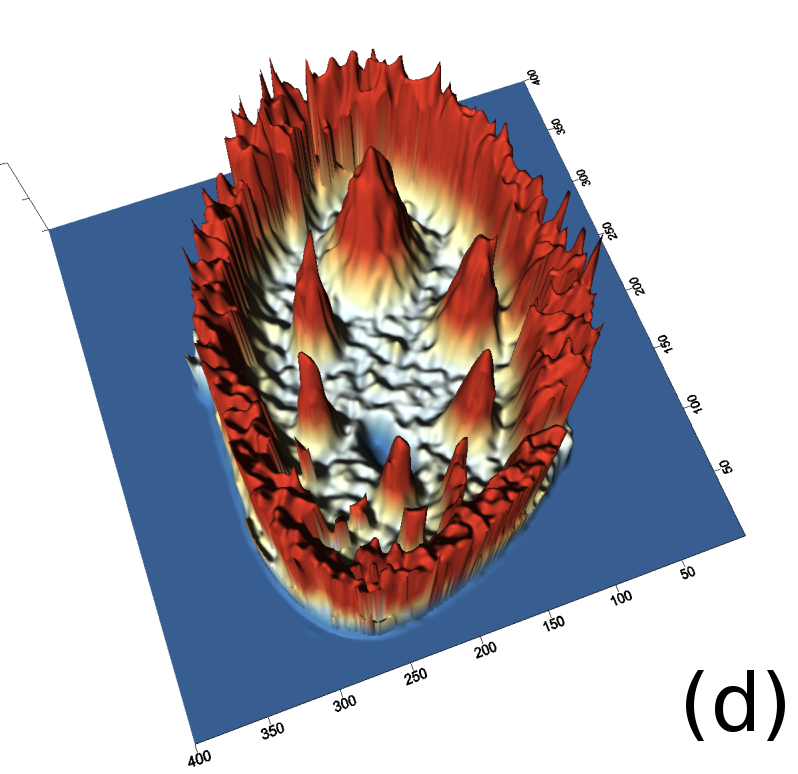}} 
  \caption{Surface representations of the reconstructed phantoms with (a) MLEM, (b) MLEM-TV, (c) MLEM-TV-$\ell_{2}$ and (d) MLEM-EL method. }
  \label{ET_rec2}
\end{figure}

In Fig. \ref{ET_rec1} and \ref{ET_rec2} we demonstrate reconstructed images for the best selected RMSE values.

In Fig. \ref{ET_rec1} and \ref{ET_rec2} one can notice that the BR is very smooth for TV and TV-$\ell_{2}$ penalties and some long-wave oscillations can be seen in the reconstructed image with EL penalty. This result corresponds to the expected behaviour of the EL penalty. We note here that the phantoms background  (see Fig. \ref{ET_phant}) is not as flat as TV and TV-$\ell_{2}$ penalty recovered it. Furthermore, a small size dot-like feature (approximately in the centre of the phantom) is almost smoothed out with TV and TV-$\ell_{2}$ recovery. However, it is visible and well recovered with EL penalty. The sharp features, overall, are reconstructed very well with MLEM-EL method and seem even sharper compare to other methods (see the bone outer rim in Fig \ref{ET_rec1}).

\section{Discussion}\label{sec:discussion}
In this paper, we emphasize that piecewise-smooth images are more favourable than piecewise-constant ones, this, however might not be the case for all reconstructed objects. Therefore, some prior knowledge about the investigated object is needed to choose an appropriate regularization term. For instance, the activity distribution in ET is smooth, therefore, penalties like EL are particularly suitable. The proposed penalty gives more gentle approach to regularization in uniform areas and does not penalize small perturbations aggressively. Moreover, the proposed method can recover small features successfully (e.g. lesions in ET case) with both step or ramp intensity variations. 

In terms of the choice of parameters, complexity of computer implementation and the speed of computation, our method is very similar to reconstruction techniques with TV penalty. Our future work will be to explore further the  space of parameters and give some recommendations for automated choice of some parameters. Additionally we will look into the issue of low-frequency oscillations of our penalty in the uniform areas. 

\section{Conclusion}\label{sec:conclusion}
In this paper, we presented a novel two-step iterative reconstruction algorithm with high-order regularization penalty. Our method outperforms in terms of signal-to-noise ratio the conventional total variation based reconstruction techniques and it is competitive with other state-of-the-art high-order based approaches. From the preliminary experiments, the proposed method is well suited for the limited  data problems in X-ray tomography as well as emission tomography.

\ack
This work has been supported by the Engineering and Physical Sciences Research Council under grants EP/J010553/1, EP/J010456/1 and EP/I02249X/1.

\end{document}